\documentclass[runningheads]{llncs}
\usepackage{graphicx}
\usepackage{amsmath,amssymb} 
\usepackage{color,verbatim,booktabs}
\usepackage[width=122mm,left=12mm,paperwidth=146mm,height=193mm,top=12mm,paperheight=217mm]{geometry}
\usepackage{wrapfig}
\usepackage{subfig}
\usepackage[final]{pdfpages}
\usepackage{hyperref}

\usepackage{xcolor}
\hypersetup{
    colorlinks,
    linkcolor={red!50!black},
    citecolor={blue!50!black},
    urlcolor={blue!80!black}
}

\newcommand{\etal}[0]{\textit{et al.\ }}

\usepackage[width=122mm,left=12mm,paperwidth=146mm,height=193mm,top=12mm,paperheight=217mm]{geometry}
\begin{document}
\pagestyle{headings}
\mainmatter

\title{Learning a Predictable and Generative \\
Vector Representation for Objects}

\titlerunning{Learning a Predictable and Generative Vector Representation for Objects}

\authorrunning{R.\ Girdhar, D.\ F.\ Fouhey, M.\ Rodriguez and A.\ Gupta}

\author{Rohit Girdhar\textsuperscript{$1$} \enskip David F. Fouhey\textsuperscript{$1$} \enskip Mikel Rodriguez\textsuperscript{$2$} \enskip Abhinav Gupta\textsuperscript{$1$}}

\institute{\textsuperscript{$1$}Robotics Institute, Carnegie Mellon University \quad
  \textsuperscript{$2$}MITRE Corporation \\
  \email { \{rgirdhar,dfouhey,abhinavg\}@cs.cmu.edu, mikel@cs.ucf.edu}
}

\maketitle

\begin{abstract}

What is a good vector representation of an object? We believe that it should be
generative in 3D, in the sense that it can produce new 3D objects;
as well as
be predictable from 2D, in the sense that it can be perceived from 2D images. We propose a
novel architecture, called the TL-embedding network, to learn an
embedding space with these properties.  The network consists of two components: (a) an autoencoder that ensures
the representation is generative; and (b) a convolutional network that ensures
the representation is predictable. This enables tackling a number of tasks
including voxel prediction from 2D images and 3D model retrieval.
Extensive experimental analysis
demonstrates the usefulness and versatility of this embedding.

\end{abstract}

\section{Introduction}

What is a good vector representation for objects? On the one hand, there has
been a great deal of work on discriminative models such as
ConvNets~\cite{AlexNet,Simonyan14c} mapping 2D pixels to semantic labels.
This approach, while useful for 
distinguishing between classes given an image, has
two major shortcomings: the learned representations do not necessarily incorporate 
the 3D properties of the objects and none of the approaches have shown strong 
generative capabilities. On the other hand, there is an alternate line of work focusing on 
learning to generate objects using 3D CAD models and deconvolutional networks \cite{chair-brox,dcign}.
In contrast to the purely discriminative paradigm, these approaches explicitly
address the 3D nature of objects and have shown success in generative tasks;
however, they offer no guarantees that their representations can be inferred
from images and accordingly have not been shown to be useful for natural image tasks.
In this paper, we propose to unify these two threads of research together and propose a new vector
representation (embedding) of objects.

\begin{figure}
\includegraphics[width=\linewidth]{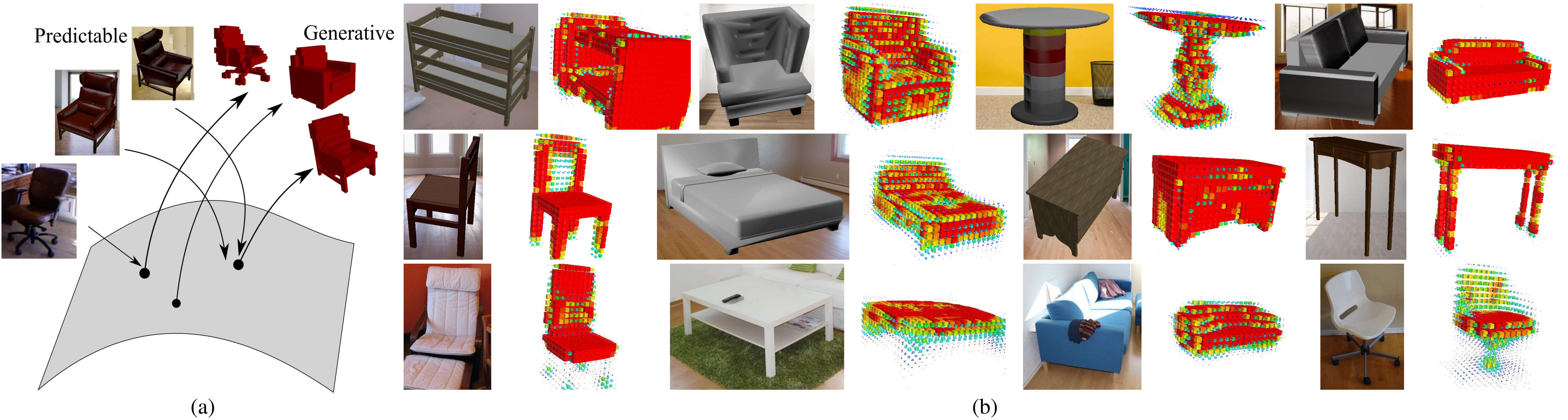}
\caption{(a) We learn an embedding space that has generative capabilities to construct 3D structures,
while being predictable from RGB images. (b) Our final model's 3D reconstruction results on natural and synthetic
test images.}
\end{figure}

We believe that an object representation must satisfy two criteria. 
Firstly, it must be {\bf generative in 3D}:
we should be
able to reconstruct 
objects in 3D from it. Secondly, 
it must be
{\bf predictable from 2D}:
we should be able to
easily infer this representation
from images. These criteria are often at odds with each other: modeling occluded voxels
in 3D is useful for generating objects but 
very difficult to predict from an image. Thus,
optimizing for only one criterion, as in most past work, tends not to obtain the other. In
contrast, we propose a novel architecture, the TL-embedding network, that
directly optimizes for {\it both} criteria. We achieve this by building an architecture
that has two major components, joined via a 64-dimensional (64D) vector embedding space:
(1) An autoencoder network which maps a 3D voxel grid to 
the 64D embedding space, and decodes it back to a voxel grid; and
(2) A discriminatively trained ConvNet 
that maps a 2D image to the 64D embedding space.
By themselves, these represent generative and predictable criteria; by joining them,
we can learn a representation that optimizes both.

At training time, we take the 3D voxel map of a CAD model 
as well as its 2D
rendered image and jointly optimize the components. The auto-encoder aims to
reconstruct the voxel grid and the ConvNet aims to predict the intermediate
embedding.  
The TL-network can be thought of as a 3D auto-encoder that tries to
ensure that the 3D representation can be predicted from a 2D rendered image. 
At test 
time, we can use the autoencoder and the ConvNet
to obtain a representation 
for 3D voxels and images respectively in the common latent space.
This enables us to tackle a variety of tasks at the
intersection of 2D and 3D.

We demonstrate the nature of our learned embedding in a series of experiments
on both CAD model data and natural images gathered in-the-wild.
Our experiments demonstrate that: (1) our representation is indeed generative in 3D, permitting
reconstruction of novel CAD models; (2) our representation is predictable from 2D, allowing us
to predict the full 3D voxels of an object from an image
(an extremely difficult task),
as well as do fast CAD model retrieval from a natural image; and (3) that the learned space 
has a number of good properties, such as being smooth, carrying class-discriminative
information, and allowing vector arithmetic. In the process,
we show the importance
of our design decisions, and the value of joining the generative and predictive approaches.

\section{Related Work}

Our work aims to produce a representation that is generative in 3D and
predictable from 2D and thus touches on two long-standing and important
questions in computer vision: how do we represent 3D objects in a vector space
and how do we recognize this representation in images? 

Learning an embedding, or vector representation 
of visual objects 
is a well studied problem 
in computer vision. In the seminal work of Olshausen and
Field~\cite{Olshausen}, the objective was to obtain 
a representation that was sparse
and could reconstruct the pixels. Since then, 
there has been a 
lot of
work in this reconstructive vein. 
For a long time, researchers focused on 
techniques such as stacked RBMs or autoencoders
\cite{HintonSalakhutdinov2006b,Vincent:2010} 
or DBMs \cite{SalHinton07}, and
more recently, this has taken the form of generative adversarial models \cite{gen-adv-nets}.
This line of work, however, has focused on building a 2D generative model of the pixels themselves. 
In this case, if the representation captures any 3D properties, 
it is modeled implicitly.
In contrast, we focus on explicitly 
modeling the 3D shape of the world. Thus, our work is most similar to 
a number of recent exceptions to the 2D end-to-end approach.
Dosovitskiy \etal\cite{chair-brox} used 3D CAD
models to learn a parameterized generative model for objects and
Kulkarni \etal\cite{dcign} introduced a
technique to guide the latent representation of a generative model to
explicitly model certain 3D properties. While they use 3D data like our work, they use
it to build a generative model for 2D images. Our work is complementary:
their work can generate the pixels for a chair and ours
can generate the voxels (and thus, 
help an agent or robot to interact with it).

There has been comparatively less work in the 3D generative space.
Past works have
used part-based models \cite{Chaudhuri:2011,Kalogerakis:2012:ShapeSynthesis}
and deep networks~\cite{3dshapenet,yangyan_fpnn_arxiv16,maturana_iros_2015}
for representing 3D models.
In contrast to 2D generative
models, these approaches acknowledges the 3D structure of the world. However, unlike
our work, it does not address the mapping from images to this 3D structure. We believe this is a crucial
distinction: while the world is 3D, the images we receive are 
intrinsically 2D and we must build our representations with this in mind.

The task of inferring 3D properties from images goes back to the very
beginning of vision. Learning-based techniques started gaining traction
in the mid-2000s \cite{Hoiem-IJCV07,Saxena08} by framing it as a supervised
problem of mapping images of scenes to 2.5D maps. Among a large body of
works trying to infer 3D representations from images, our approach is
most related to a group of works using renderings of 3D CAD models to 
predict properties such as object viewpoint~\cite{Su_2015_ICCV} or
class~\cite{su15mvcnn}, among others~\cite{shapeFrom3D,object_pose_in_rgbd,PengSAS14}.
Typically, these approaches focus on global 3D properties such as pose in the
case of objects, and 2.5D maps in the case of scenes.
Our work predicts a much more challenging representation, a voxel map 
(i.e., including the occluded parts). 
Related works in 3D prediction include \cite{Kar15,wu2016single,choy_3dr2n2_arxiv16}.
Our approach differs from these as it is class agnostic, voxel based and learns
a joint embedding that enables various applications beyond 3D prediction.

Our final output is related to CAD model retrieval in the sense
that one output of our approach is a 3D model. Many approaches
achieve this via alignment~\cite{Lim13,Aubry14,Huang15} or
joint, but non-generative embeddings \cite{Li15}.
In contrast
to these works, we take the extreme approach of generating the
3D voxel map from the image. While we obtain coarser
results than using an existing model, this explict generative mapping gives
the potential to generalize to previously unseen objects.

\section{Our Approach}

\begin{figure}[t]
\centering
\includegraphics[width=0.95\linewidth]{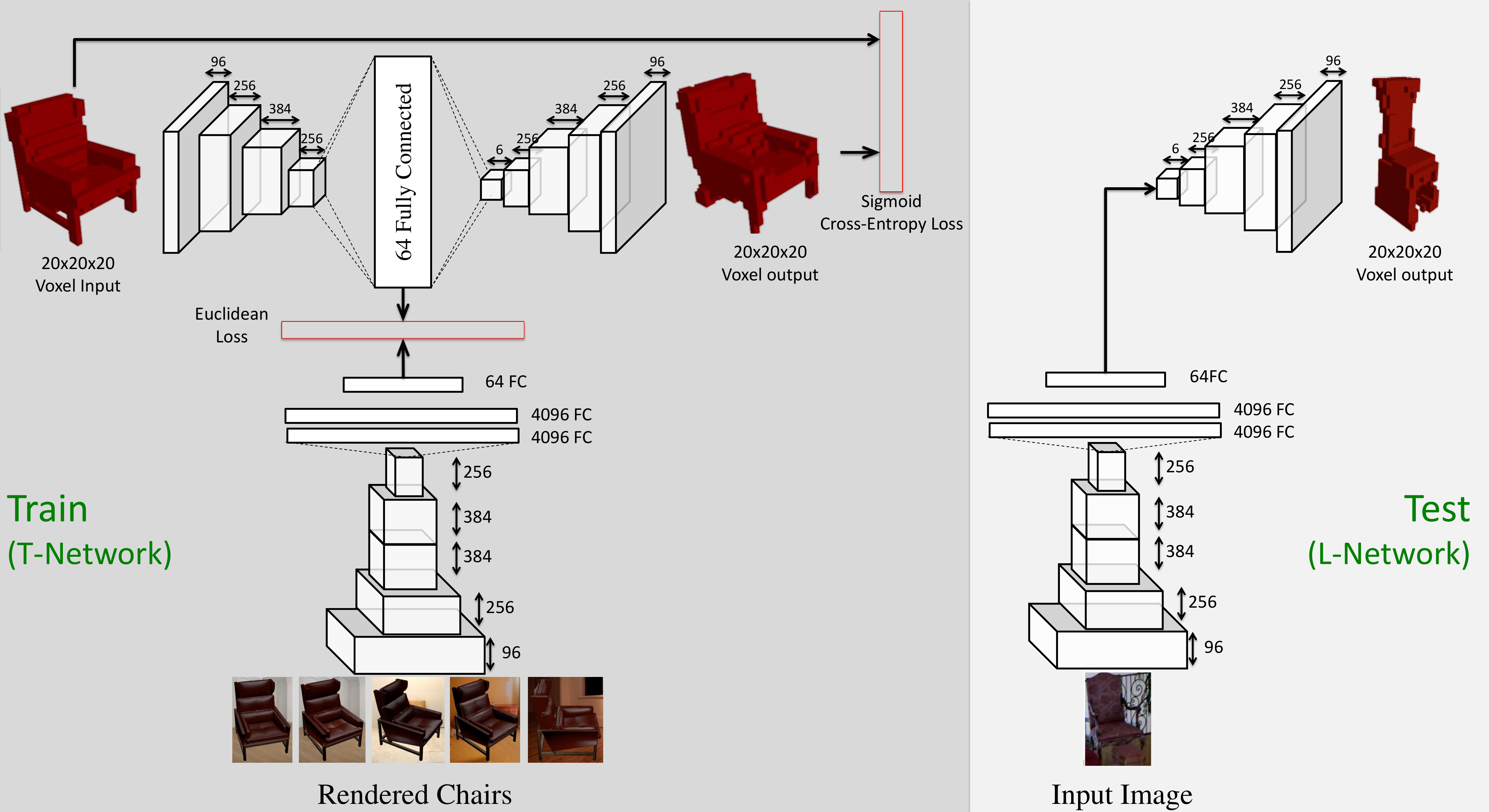}
\caption{Our proposed TL-embedding network. 
{\bf (a) T-network:} At training time, the network takes two inputs:
2D RGB images which are fed into ConvNet at the bottom and 3D voxel maps
which are fed into the autoencoder on the left. The output is a 3D voxel
map.  We apply two losses jointly: a reconstruction loss for the voxel
outputs, and a regression loss for the 64-D embedding in the middle. 
{\bf (b) L-network:} During testing, we remove the encoder part and
only use the image as input. The ConvNet predicts the embedding
representation and the decoder predicts the voxel.}
\label{fig:nwarch}
\end{figure}

To reiterate, our goal is to 
learn a vector representation that is:
(a) {\bf generative:} we should be able to generate voxels in 3D from 
this representation; and
(b) {\bf predictable:} we should be able to take a 2D image of an object
and predict this representation. Both properties are vital for image
understanding tasks.

We propose a novel TL-embedding network (Fig.~\ref{fig:nwarch}) 
to optimize both these criteria. The T and L refer to the architecture
in the training and testing phase. The top part of the T network is an
autoencoder with convolution and deconvolution layers. The encoder maps 
the 3D voxel map to a low-dimensional subspace. The decoder maps a
datapoint in the low-dimensional subspace to a 3D voxel map. The 
autoencoder forces the embedding to be generative, and we can
sample datapoints in this embedding to reconstruct new objects.
To optimize the predictable criterion, we use a ConvNet
architecture similar to AlexNet~\cite{AlexNet}, adding a
loss function that ensures the embedding space is predictable
from pixels.

Training this TL-embedding network requires 2D RGB images and their corresponding 3D voxel maps.
Since this data is hard to obtain, we use CAD model datasets to obtain voxel
maps and render these CAD models with different random backgrounds to generate
corresponding image data. We now describe our network architecture and the
details of our training and testing procedure.

\noindent {\bf Autoencoder Network Architecture:}
The autoencoder takes a $20\times20\times20$ 
voxel grid representation
of the CAD model as input. The encoder consists of four convolutional layers
followed by a fully connected layer that produces an embedding vector.
The decoder takes this embedding and maps it to a 
$20^3$ voxel grid with five deconvolutional layers. Throughout, we
use 3D convolutions with stride 1, connected via parameterized
ReLU \cite{he2015delving} non-linearities.

We train the autoencoder with a Cross-Entropy loss on the
final voxel output against the original voxel input. This loss function has
the form:
\begin{eqnarray}\label{eq:loss}
E = -\frac{1}{N} \sum_{n=1}^{N} [p_n \log \hat{p}_n + (1 - p_n) \log(1 - \hat{p}_n)]
\end{eqnarray}
where $p_n$ is the target
probability (1 or 0) of a voxel being filled, $\hat{p}_n$ is the predicted probability obtained through
a sigmoid, and $N = 20^3$.

\noindent {\bf Mapping 2D Image to Embedding Space:} 
The lower part of the T
network learns a mapping from 2D image space to the 64D embedding
space. We adopt the AlexNet architecture~\cite{AlexNet} which has five
convolutional layers and two fully connected layers. We add
a 64D fc8 layer to the original AlexNet architecture and use a
Euclidean loss. We initialize this network with the parameters
trained on ImageNet \cite{deng2009imagenet} classification task.

One strength of our TL-embedding network is that it can be used to
predict a 3D voxel map for a given 2D image. 
At test time, we remove
the encoder part of the autoencoder 
network and connect the output of the image
embedding network to the decoder to obtain this voxel output.

\subsection{Training the TL-Embedding Network}

We train the network using batches of (image, voxel) pairs. The images are
generated by rendering the 3D model and the network is then trained in a three
stage procedure. We now describe this in detail.

\begin{figure}[t]
\centering
\includegraphics[trim={0 12cm 0 0},clip,width=\linewidth]{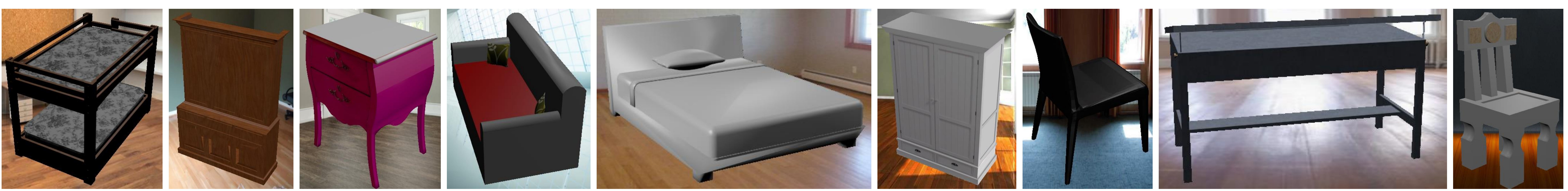}
\caption{Sample renderings used to train 
our network.
We render each training model
into 72 views
over a random background each epoch of training.}
\label{fig:render-egs}
\end{figure}

\noindent {\bf Data Generation:}
We use ideas from~\cite{Su_2015_ICCV} to render the 3D models for training our
network. To prevent the network from overfitting to sharp edges when rendered
on a plain background, we render it on randomly selected open room images
downloaded from the internet. Following the popular practice~\cite{su15mvcnn},
we render all the models into 72 views, at
three elevations of 15, 30 and 45
degrees and 24 azimuth angles from 0 to 360 degrees, in increments of
15 degrees. We convert the 3D models into $20^3$ voxel
grid using the voxelizer from~\cite{3dshapenet}.

\noindent {\bf Three-stage Training:} 
Training a TL-embedding network from scratch
and jointly is a challenging problem. 
Therefore, we take a three stage procedure.
(1) In the first stage, we train the autoencoder part
of the network independently. This network is
initialized at random, and trained end-to-end with the sigmoid cross-entropy
loss. We train this for about 200 epochs. 
(2) In the second stage we train the ConvNet to regress to the 64D representation.
Specifically, the encoder generates the embedding for the voxel and the image
network is trained to regress the embedding.
The image network is initialized using ImageNet pre-trained weights.
We keep the lower convolutional layers fixed.
(3) In the final stage, we finetune the network jointly
with both the losses.
In this stage, we observe that the prediction loss reduces significantly while reconstruction
loss reduces marginally. We also observe that most of 
the parameter update happens in the autoencoder network, indicating that the autoencoder
updates its latent representation to make it easily predictable from images, while maintaining or
improving the reconstruction performance given this new latent representation.

\noindent {\bf Implementation Details:}
We implement this network using the Caffe~\cite{Jia2014caffe} toolbox. In the first stage,
we initialize all layers of autoencoder
network from scratch using $\mathcal{N}(0,0.01)$ and train with a uniform learning rate of $10^{-6}$.
Next, we train the image network by initializing fc8 from scratch and remaining layers from
ImageNet. We finetune all layers after and including conv4 with a uniform learning rate of $10^{-8}$. 
A lower learning rate is required because the initial prediction loss values are in the range of 500K.
The encoder network from the autoencoder is used in testing-phase with its previously learned weights
to generate the labels for image network.
Finally, we jointly train using both losses, initializing the network using weights learned earlier,
and finetuning all layers of autoencoder and all layers after and including conv4 for image network
with a learning rate of $10^{-10}$. 
Since our network now has two losses, we balance their values
by scaling the autoencoder loss to have approximately same initial value, 
as otherwise the network tends to optimize for 
the prediction loss without regard to the reconstruction loss.

\section{Experiments}

\begin{figure}[t]
\includegraphics[width=\textwidth]{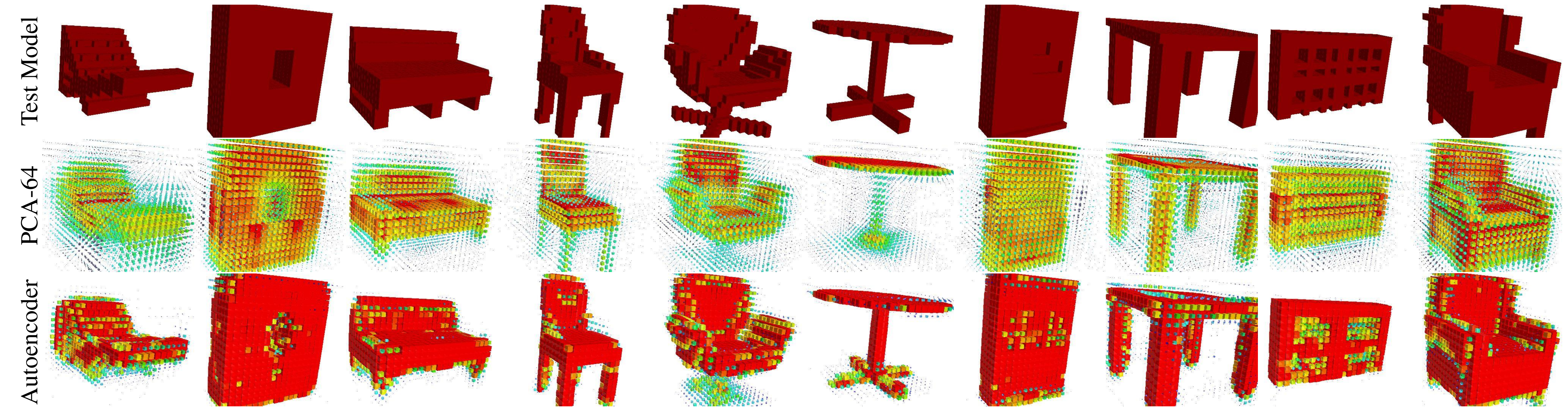}
\caption{Reconstructions of {\it random} test models using PCA and the autoencoder.
Predicted voxels are colored and sized by confidence of prediction, 
from large and red to small and blue in decreasing order of confidence.
PCA is much less confident about the extent as well as fine details
as compared to our autoencoder.}
\label{fig:reconstruction}
\end{figure}

\begin{table}[t]
\centering
\caption{Reconstruction performance using AP on test data.}
\label{tab:reconstruction}
\begin{tabular}{r@{~~~}ccccc@{~~~}c} \toprule
                          & Chair         & Table         & Sofa          & Cabinet       & Bed           & Overall \\ \midrule
Proposed (before Joint)   & \bf 96.4    & \bf 97.1    & 99.1    & \bf 99.3    & \bf 94.1    & \bf 97.6    \\
Proposed (after Joint)    & \bf 96.4    & 97.0    & \bf 99.2    & \bf 99.3    & 93.8    & \bf 97.6    \\
PCA                       & 94.8        & 96.7        & 98.6        & 99.0        & 91.5        & 96.8 \\ \bottomrule

\end{tabular}
\end{table}

We now experimentally evaluate the method. 
Our overarching goal is to answer the following questions: (1) is the
representation we learn generative in 3D? (2) can the representation be
predicted from images in 2D? In addition to directly answering these questions,
we verify that the 
model has learned a sensible latent representation by ensuring
that the latent representation satisfies a number of properties, such as
being smooth, discriminative and allowing arithmetic.

We note that our approach
has a capability that, to the best of our knowledge, is previous unexplored: it can
simultaneously reconstruct in 3D and predict from 2D. Thus, there are no standard baselines or
datasets for this task. Instead, we adopt standard datasets for each of the many tasks
that our model can perform. Where appropriate, we compare the method with existing methods.
These baselines, however, are specialized solutions to only one of the many tasks we can solve
and often use additional supervisory information.
As the community starts tackling increasingly difficult 3D problems like direct voxel
prediction, we believe that our work can be a strong baseline to benchmark progress.

We proceed as follows. We introduce the datasets and evaluation criterion that we use in Sec.\ \ref{sec:exp_data}.
We first verify that our learned representation models the space
of voxels well in a number of ways: that it is reconstructive, smooth, and can be used
to distinguish different classes of objects (Sec.\ \ref{sec:exp_reconstruction}). This evaluates the representation
independently of its ability to predict voxels from images. We then  verify that
our approach can predict the voxels from 2D and show that it outperforms alternate options (Sec.\ \ref{sec:exp_pred}). Subsequently, we show that our representation can
be used to do CAD retrieval from natural images (Sec.\ \ref{sec:cad_retrieval}) and is capable of performing 3D shape arithmetic  (Sec.\ \ref{sec:exp_arithmetic}).

\subsection{Datasets and Evaluation}
\label{sec:exp_data}

We use two datasets for evaluation. The first is a CAD model dataset
used to train the TL-embedding and to explore the learned embedding. The second
is an in-the-wild dataset used to verify that the
approach works on natural
images.

\noindent {\bf CAD Dataset:}
We use CAD models from the ShapeNet\cite{3dshapenet} database. This database contains over
220K models organized into 3K WordNet synsets. We take a set of common indoor
objects: chair (6778 models), table (8509 models), sofa (3173 models), cabinet
(1572 models), and bed (254 models). We split these models randomly into 16228 train 
and 4058 test objects. All our models are trained with rendered images 
and voxels from the above train set. We use the test set to 
quantify our performance and analyze our models.
\par
\noindent {\bf IKEA Dataset:} 
We quantify the performance of our model on natural indoor images 
from IKEA Dataset \cite{Lim13} which are labeled with 3D models.
Since our approach expects to reconstruct a single
object, we test it on cropped images of these objects.
These boxes, however, include cluttered backgrounds and pieces of other objects.
After cropping these objects out of provided 759 images, we get 937 images labeled with one of provided 225 3D models.
\par
\noindent {\bf Evaluation Metric:} 
Throughout the paper, we use Average Precision (AP) over the complete test set
to evaluate reconstruction performance. We also show per-class APs where applicable
to better characterize our model's performance.

\subsection{Embedding Analysis}
\label{sec:exp_reconstruction}

We start by probing our learned representation in terms of 3D voxels.
Here, we focus on the autoencoder part of the network -- that is, we feed a voxel grid to the network
and verify a number of properties: (a) that it can reconstruct the voxels well qualitatively
and quantitatively, which verifies that the method works; (b) that it outperforms
a linear baseline, PCA, for reconstruction, which further validates the choice of a convolutional
autoencoder; and (c) that the learned representation
is smooth and carries class-discriminative
information, which acts as additional 
confirmation that the representation is meaningful.

\noindent {\bf Qualitative Results:} 
First, we show qualitative results: Fig.\  \ref{fig:reconstruction} shows randomly selected reconstructions using the autoencoder and PCA. While a simple
linear approach is sufficient to capture the coarse structure, our approach
does much better at fine-details (chair legs in col. 6) as well as at getting the extent
correct (back of the chair in col. 4). Note also the large amount of low but
non-zero probability voxels in the free-space in PCA compared to the
auto-encoder.

\begin{figure}[t]
\centering
\subfloat[]{\includegraphics[width=0.54\textwidth]{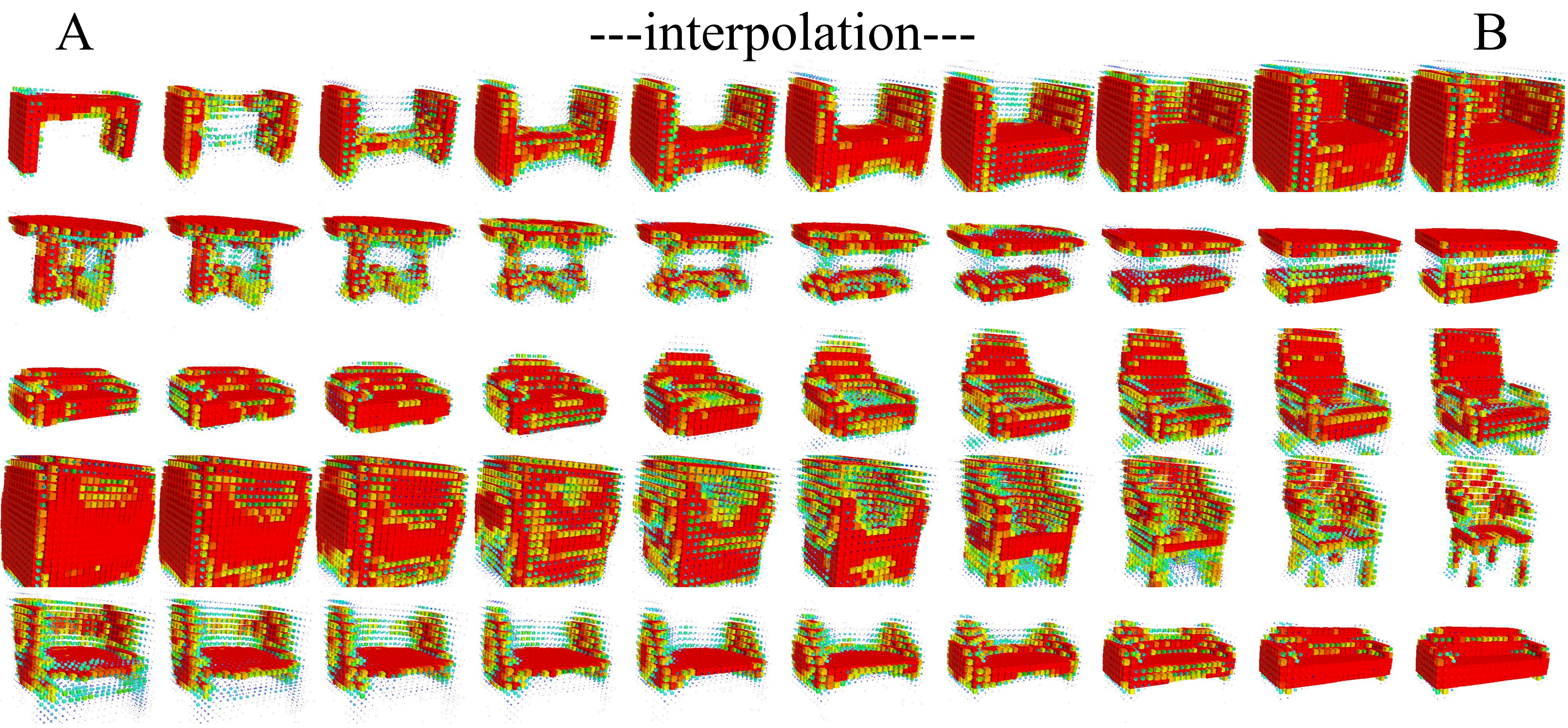}} \hfill
\subfloat[]{\includegraphics[width=0.44\textwidth]{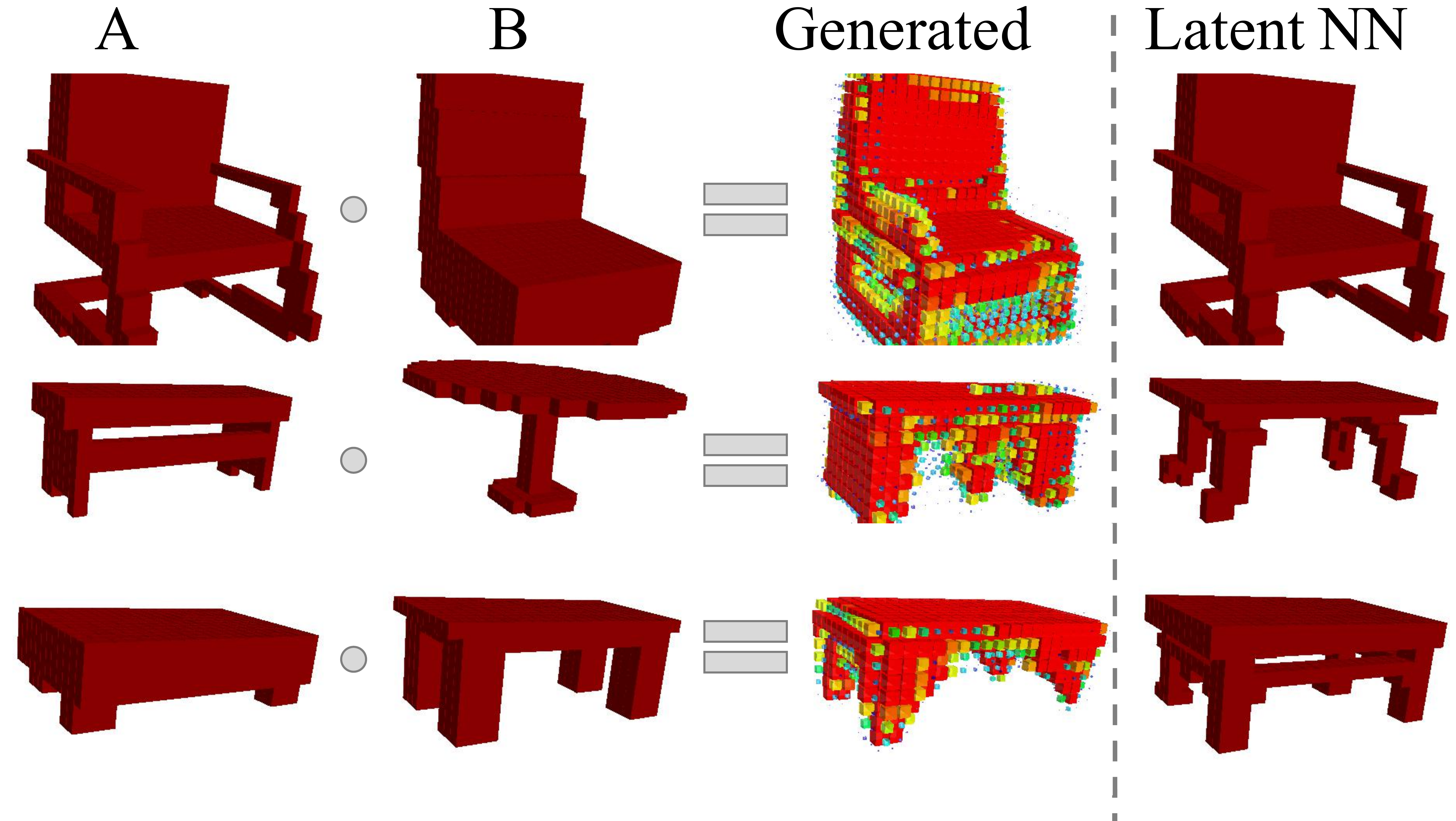}}
\caption{(a) Reconstructions for linear interpolation between
two randomly picked latent representations. (b) Evaluating generative ability by combining dimensions from two 
training models. We show the reconstruction and the nearest neighbor in the
training set (over latent features). The difference shows we can generate novel
models, such as an armchair with one arm-rest.}
\label{fig:randomsample}
\end{figure}

We next show that the learned space is smooth, by computing reconstructions for
linear interpolation between latent representations of randomly picked test models. 
As Fig.\ \ref{fig:randomsample}(a) shows,
the 3D models smoothly transition in structure and most intermediate models are
also physically plausible.
We also show results exploring the learned space and verifying whether 
the dimensions are meaningful. One way to do this is to generate 
new points 
in the space and reconstruct them. We generate
these points by taking the first 32 dimensions from one model and the rest from
another. As seen by the difference between the reconstruction
and the nearest model in Fig.\ \ref{fig:randomsample}(b), this can generate previously
unseen models that combine aspects of each model.

\begin{figure}[t]
\centering
\includegraphics[width=\textwidth]{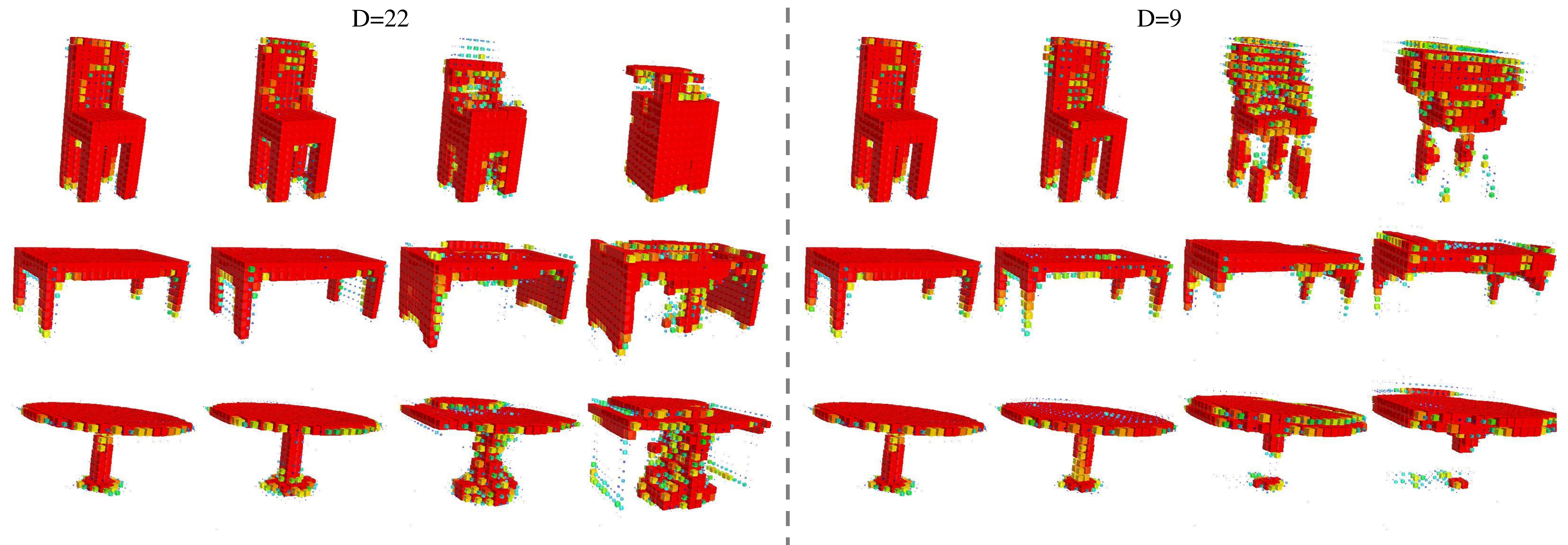}
\caption{We evaluate if the dimensions are meaningful by scaling each dimension
separately and analyzing the effect on the reconstruction. Some dimensions have 
a consistent effect on reconstruction across objects. Higher values
in dimension 22 lead to thicker legs, and higher values in 9 lead to disappearance of legs.}
\label{fig:dimscale}
\end{figure}

We further attempt to understand the embedding space by
clamping all the dimensions of a latent vector but one and scaling the 
selected dimension by adding a fixed value to it.
We show its effect on two dimensions and three models
in Fig.\ \ref{fig:dimscale}. Such scaling of these
dimensions produces consistent
effects across models, suggesting that some learned dimensions are 
semantically meaningful.

\noindent {\bf Quantitative Reconstruction Accuracy:} We now evaluate the
reconstruction performance quantitatively on the CAD test data and report
results in Table \ref{tab:reconstruction}. Our goal here is to verify that the
auto-encoder is worthwhile: we thus compare to PCA using the same
number of dimensions. 
Our method obtains extremely high performance, $97.6\%$ AP and consistently outperforms PCA, reducing the average error rate by $25\%$ relative. It can be seen
in Table~\ref{tab:reconstruction} that some categories are easier than others:
sofas and cabinets are naturally more easy than beds (including bunk-beds) and chairs.
Our method consistently obtains larger gains on challenging objects, indicating the 
merits of a
non-linear representation. We also evaluate the performance of 
the autoencoder after the joint training. Even after being optimized to be more predictable
from image space, we can see that it still 
preserves the overall reconstruction performance.

\noindent {\bf CAD Classification:} If our representation models 3D well,
it should permit us to distinguish different types 
of objects. We empirically verify this by using our approach
{\it without modifications} as a representation to classify 3D shapes.
Note that while adding a classification loss and finetuning might 
further improve results, it would defeat the purpose of this
experiment, which is to see whether the model learns a good 3D representation on
its own. 
We evaluate our representation's performance for a classification task on the Princeton ModelNet40~\cite{modelnet_dataset} dataset
with standard train-test split from~\cite{3dshapenet}. 
We train the network on all 40 classes (again: no class information is provided)
and then use the autoencoder representation as a feature for
40-way classification. Since our representation is low-dimensional (64D), we
expand the feature to include pairwise features and train a linear SVM. Our approach obtains an
accuracy of $74.4\%$. This is within $2.6\%$ of
\cite{3dshapenet}, a recent approach on voxels that uses
class information at representation-learning time, and finetunes the representation
discriminatively for the classification experiment. Using a 64D PCA representation
trained on ModelNet40 trainset with the same feature augmentation and linear SVM obtains $68.4\%$.
This shows that our representation is class-discriminative
despite not being trained or designed so, and outperforms the
PCA.

\subsection{Voxel Prediction}
\label{sec:exp_pred}

\begin{figure}[t]
\includegraphics[width=\textwidth]{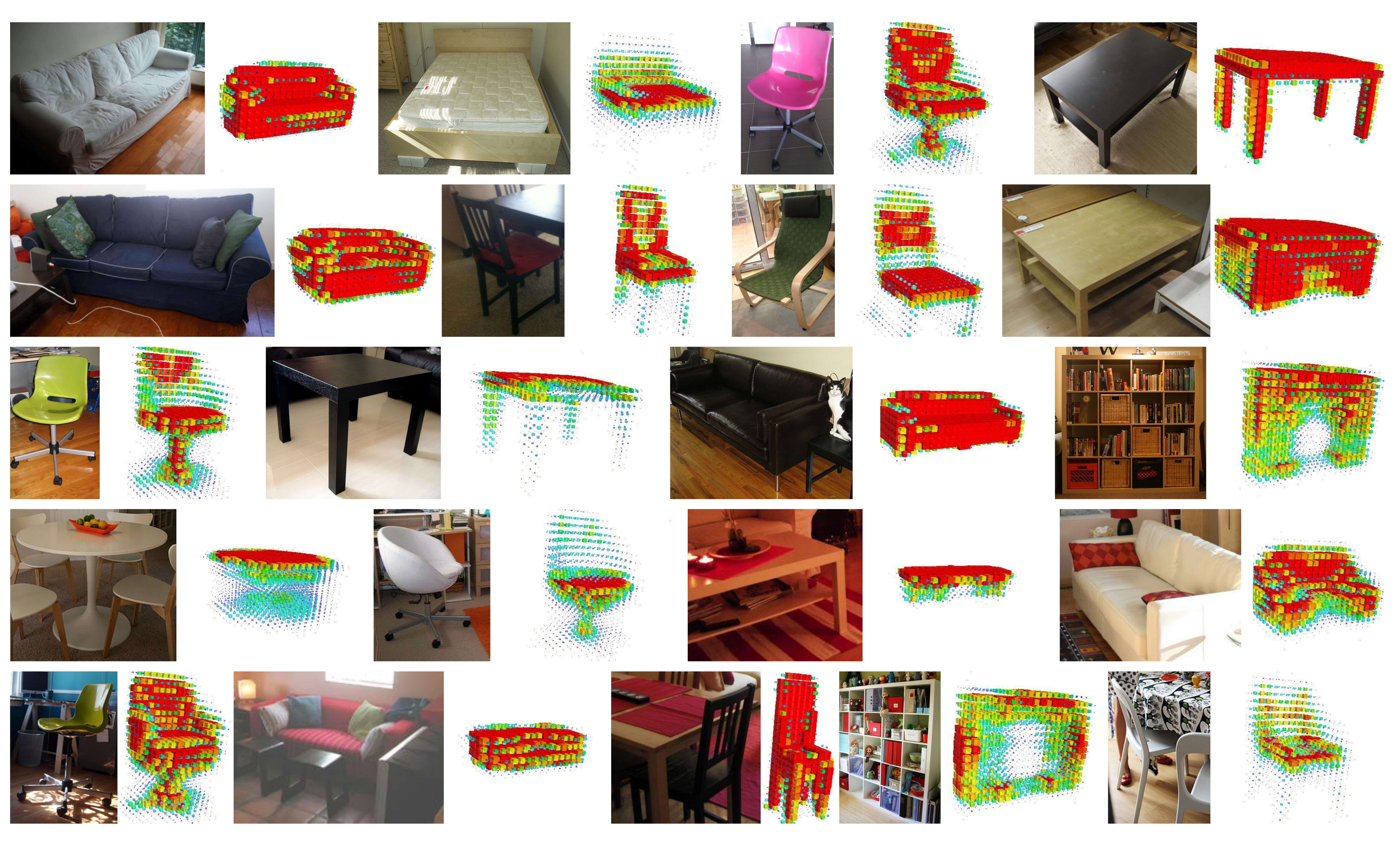}
\caption{Reconstruction results on the IKEA dataset. 
Our model generalizes well to real images, even to bookshelves which our model is not trained on. 
}
\label{fig:IKEA}
\end{figure}

We now turn to the task of predicting a 3D voxel grid from an image. We obtain
strong performance on this task and outperform a number of baselines,
demonstrating the importance of each part of our approach.

\noindent {\bf Baselines:} To the best of our knowledge, there are no 
methods that directly predict voxels from an image; we therefore compare
to a direct prediction method as well an ablation study, where we do not
perform joint training. Specifically:
(a) {\it Direct}: finetuning the ImageNet pre-trained AlexNet to predict the $20^3$ 
voxel grid directly. This corresponds to removing the auto-encoder.
We tried two strategies for freezing the layers:
Direct-conv4 refers to freezing all layers before conv4
and Direct-fc8 refers to freezing all layers except fc8.
(b) {\it Without Joint}: training the T-L network without the final
joint fine-tuning (i.e., following only the first two training stages).
The direct baselines test whether the auto-encoder's low-dimensional representation is 
necessary and the without-joint tests whether learning the model to be jointly
generative {\it and} predictable is important.

\begin{table}[t]
\centering
\caption{Average Precision for Voxel Prediction on the CAD test set. The Proposed
TL-Network outperforms the baselines on each object.}
\label{tab:CAD}
\begin{tabular}{r@{~~~}ccccc@{~~~~}c} \toprule
                          & Chair     & Table     & Sofa      & Cabinet   & Bed       & Average   \\ \midrule
Proposed (with Joint)     & \bf 66.9  & \bf 59.7  & \bf 79.3  & \bf 79.3  & \bf 41.9  & \bf 65.4  \\
Proposed (without Joint)  & 66.6      & 57.5      & \bf 79.3  & 76.5      & 33.8      & 62.7      \\
Direct-conv4              & 40.9      & 23.7      & 58.1      & 44.3      & 23.1      & 38.0      \\ 
Direct-fc8                & 21.8      & 15.5      & 35.6      & 32.7      & 18.6      & 24.8      \\ \bottomrule
\end{tabular}
\end{table}

\begin{table}[t]
\centering
\caption{Average Precision for Voxel Prediction on the IKEA dataset.}
\label{tab:ikea-quant}
\begin{tabular}{r@{~~~}cccccc@{~~~~}c} \toprule
                & Bed       & Bookcase  & Chair     & Desk      & Sofa      & Table         & Overall       \\ \midrule
Proposed        & \bf 56.3  & \bf 30.2  & \bf 32.9  & 25.8      & \bf 71.7  & \bf 23.3      & \bf 38.3      \\
Direct-conv4    & 38.2      & 26.6      & 31.4      & \bf 26.6  & 69.3      & 19.1          & 31.1          \\
Direct-fc8      & 29.5      & 17.3      & 20.4      & 19.7      & 38.8      & 16.0          & 19.8          \\ \bottomrule
\end{tabular}
\end{table}

\noindent {\bf Qualitative Results:} We first show qualitative results on natural images
in Fig.\ \ref{fig:IKEA}. Note that our method automatically predicts occluded
regions of the object, unlike most work on single image 3D (e.g.,
\cite{Saxena08,Hoiem-IJCV07,Eigen14,Eigen15,Fouhey13a,Wang15a}) that predict a
2.5D shell. For instance, our method predicts all four legs of 
furniture even if fewer are visible.  
Our model generalizes well to natural images even though 
it was trained on CAD models. Note that for instance, the round and
rectangular tables are predicted as being round and rectangular, and 
office chairs on a single post and four-legged chairs can be distinguished. One difficulty
with this data is that objects are truncated or occluded and some windows
contain multiple objects; our model does well on this data, nonetheless.

\noindent {\bf Quantitative Results:}
We now evaluate the approach quantitatively on both datasets.
We report results 
on the CAD dataset in Table \ref{tab:CAD}. Our approach outperforms
all the baselines.
Directly predicting
the voxels does substantially worse because predicting all the voxels is
a very
difficult task compared to our embedding space. Not doing joint training
produces worse results because the embedding is not forced to be predictable.

The IKEA dataset is more challenging because it is captured in-the-wild, but
our approach still produces quantitatively strong performance.
While the CAD Dataset models are represented in canonical form, the 
IKEA models are provided in no consistent orientation. We thus attempt to align each prediction with the ground-truth model by 
taking the best rigid alignment over permutations, flips and translational alignments (up to 10\%) 
of the prediction.
As Table \ref{tab:ikea-quant} shows, our approach outperforms the direct prediction 
by a large margin (38\% compared to 31\%).
If we do not correct for translational alignments, we still outperform the baseline (33\% vs 28\%).
Directly predicting voxels again performs worse compared to predicting the latent space
and reconstructing, validating the idea of using a lower-dimensional representation of
objects.

\begin{figure}[t]
\includegraphics[width=\linewidth]{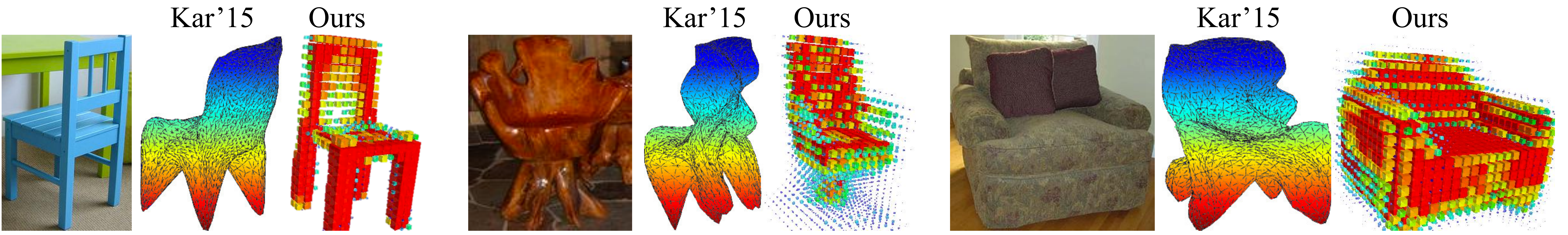} 
\caption{Predictions on PASCAL 3D+ images using \cite{Kar15} and our method.
Our method is better at capturing fine stylistic details, like the 
straight legs and the hollow back in the first case, 
a single central leg in the second, and
no visible legs in the last.}
\label{fig:kar-compare}
\end{figure}

\noindent {\bf Comparison with Kar \etal\cite{Kar15}:}
We also compare our method with~\cite{Kar15} on PASCAL 3D+
v1.0~\cite{xiang_wacv14} dataset for categories that overlap with our training categories (chair and sofa).
As Fig.\ \ref{fig:kar-compare} shows, our output is more varied and captures stylistic details better.
For quantitative comparison, we voxelize their output and 
ground truth,
and compute the overlap P-R curve with alignment.
Since \cite{Kar15} produces a binary non-probabilistic prediction
and thus yields only one operating point, we compare via maximum F-1 score
instead of AP.
After aligning, we outperform
their method 0.492 to 0.463.

\subsection{CAD Retrieval}
\label{sec:cad_retrieval}

\begin{figure}[t]
\includegraphics[width=\linewidth]{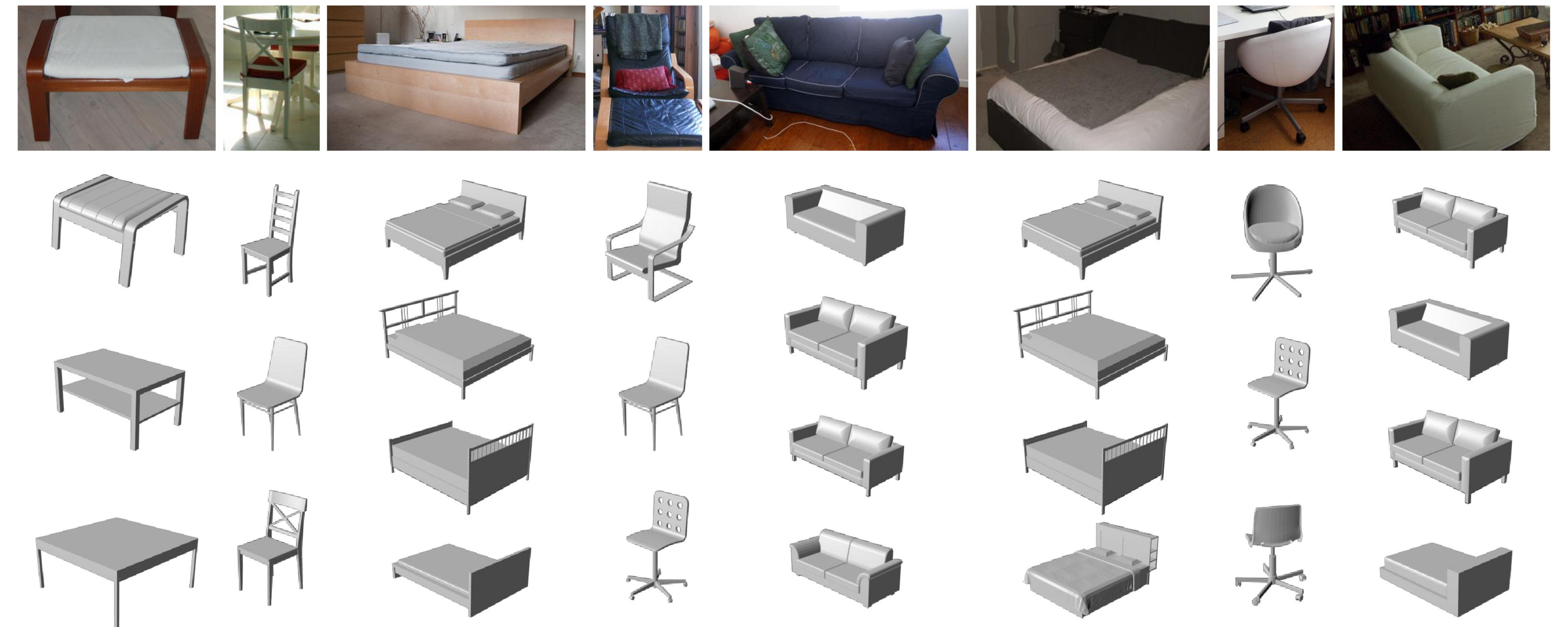} 
\caption{Top CAD model retrievals from natural images from the IKEA dataset.}
\label{fig:cadRetrieval}
\end{figure}

\begin{figure}
\centering
\begin{tabular}{ccccc}
  Chair & Bed & Sofa & Bookcase & Table/Desk \\
  \raisebox{1.3em}{\rotatebox{90}{\scriptsize{Instance}}}
	\includegraphics[width=0.19\linewidth]{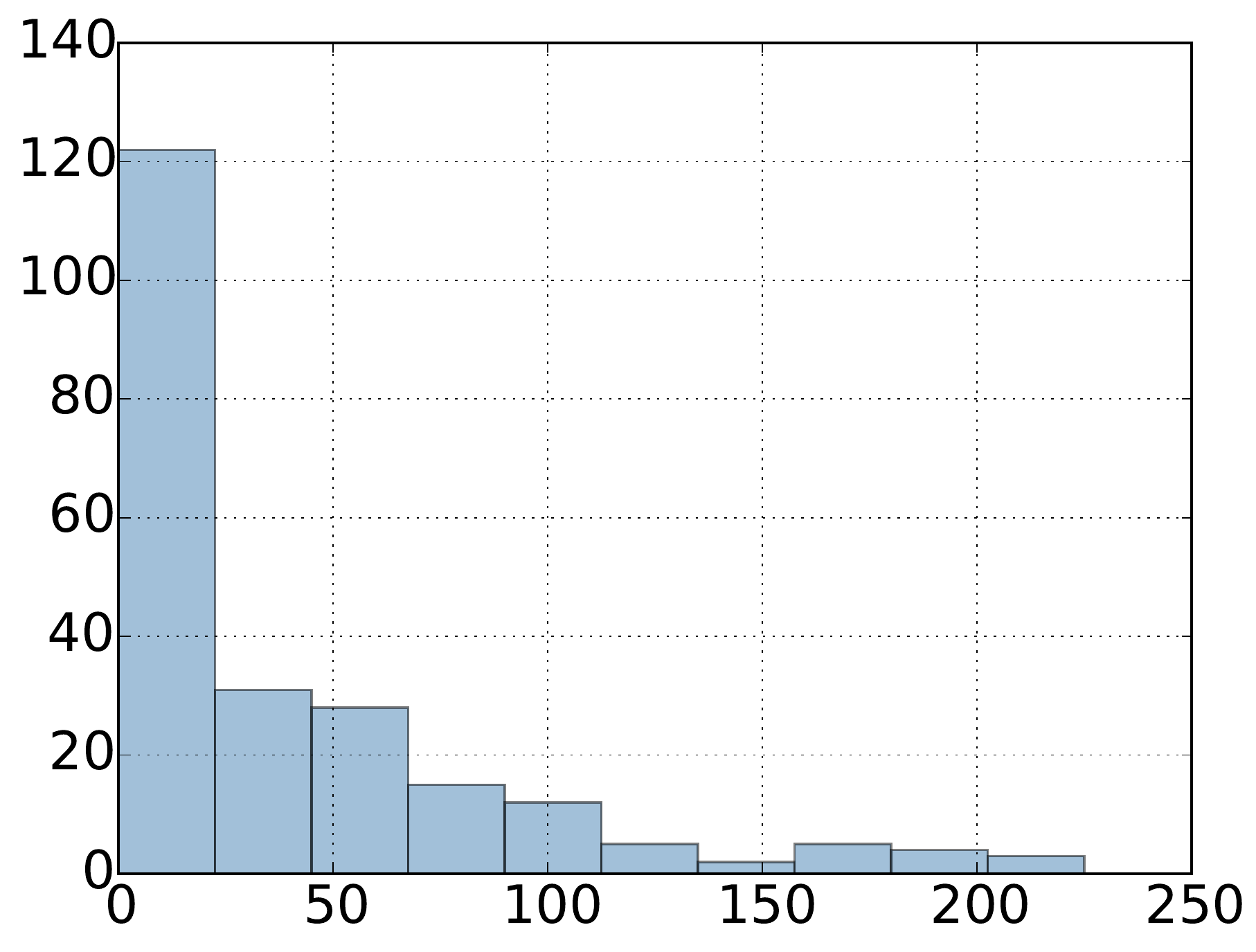} &
	\includegraphics[width=0.19\linewidth]{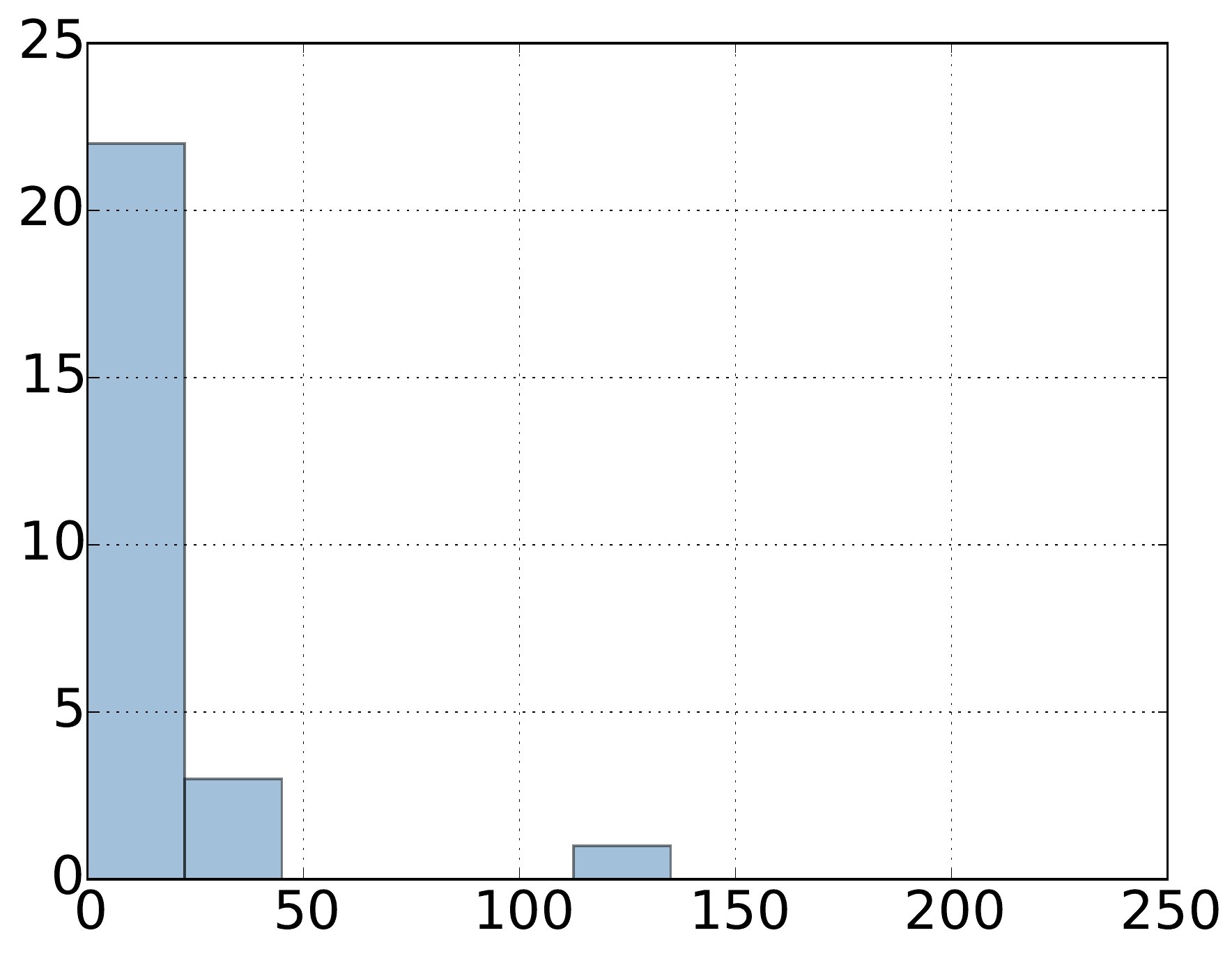} &
	\includegraphics[width=0.19\linewidth]{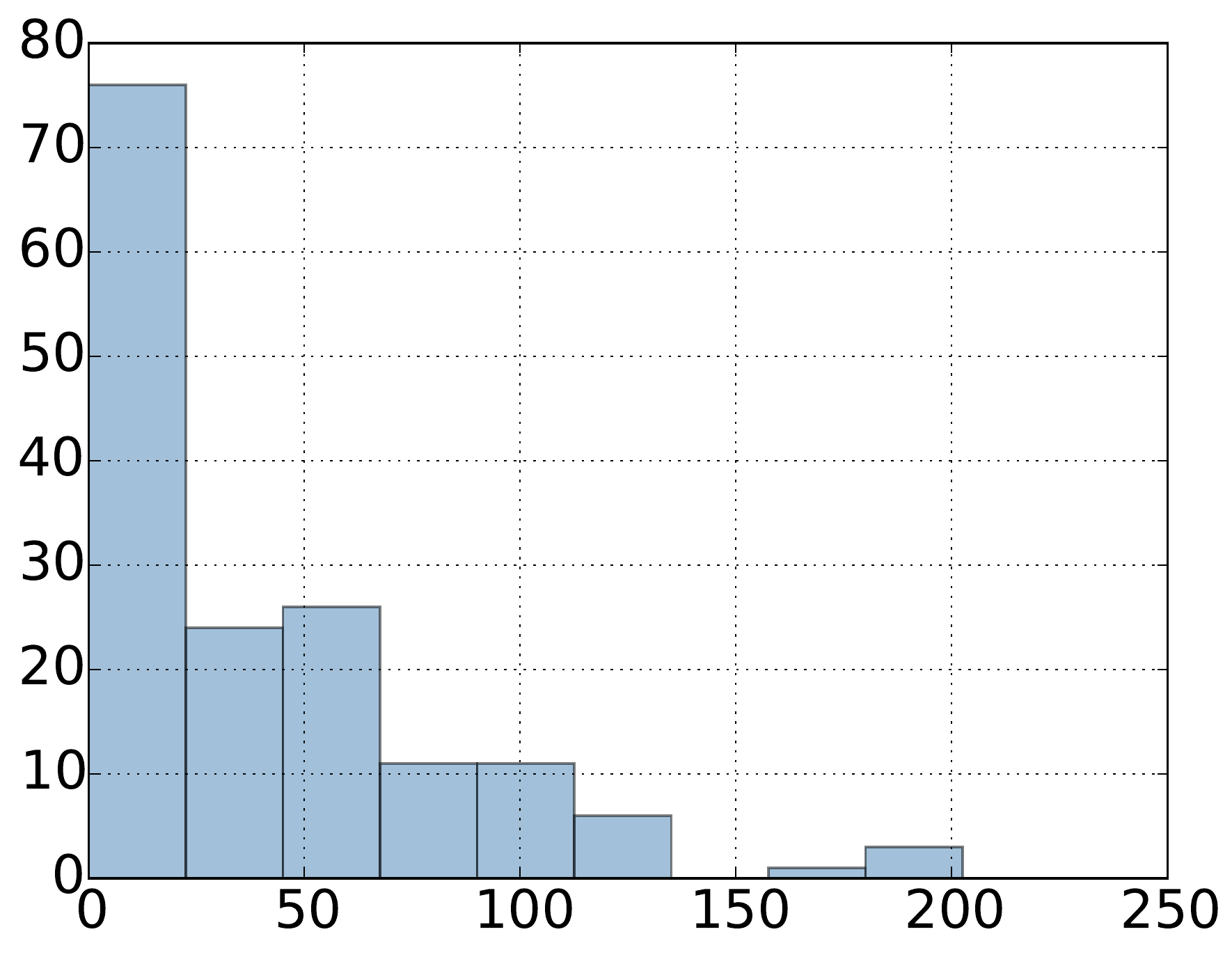} &
	\includegraphics[width=0.19\linewidth]{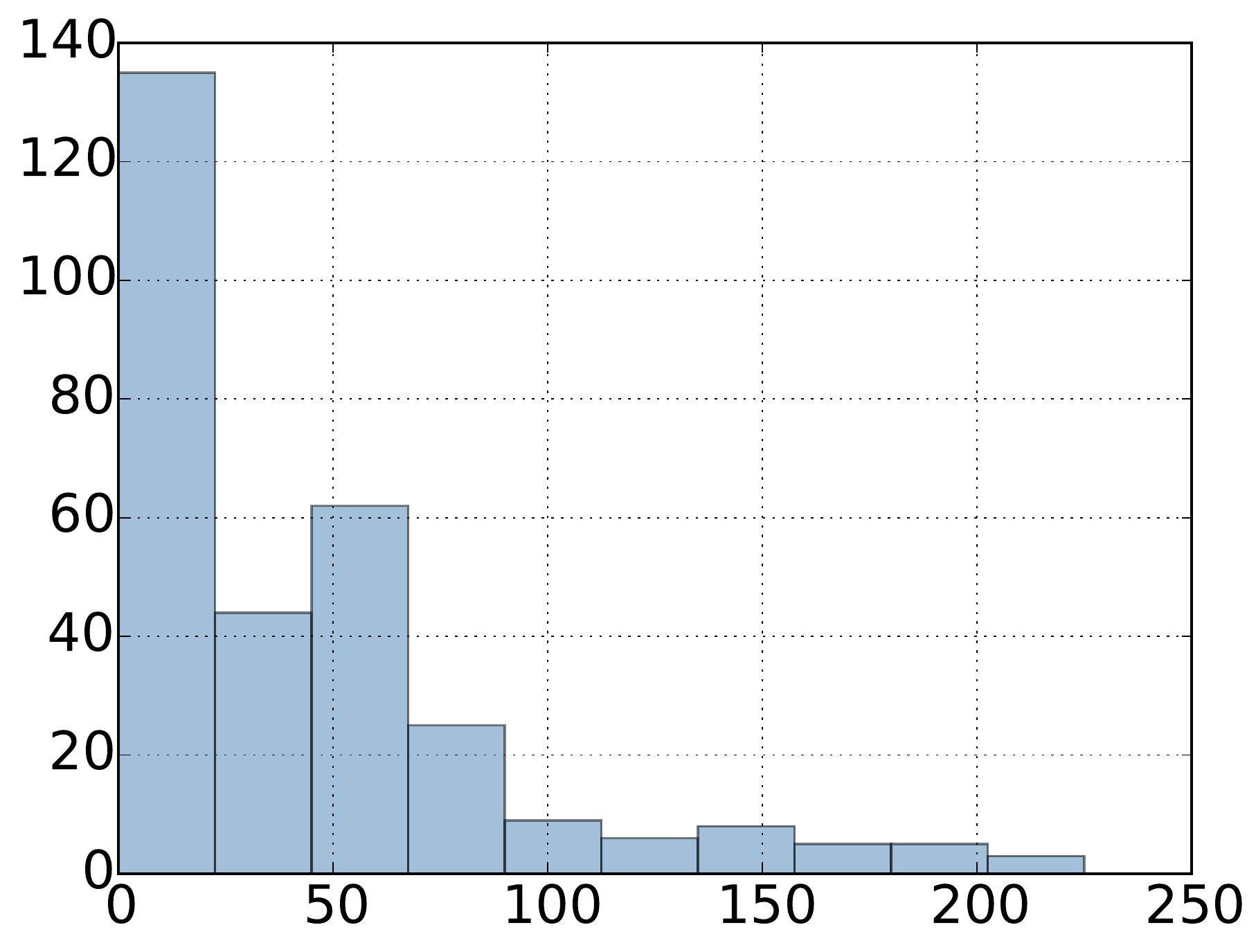} &
	\includegraphics[width=0.19\linewidth]{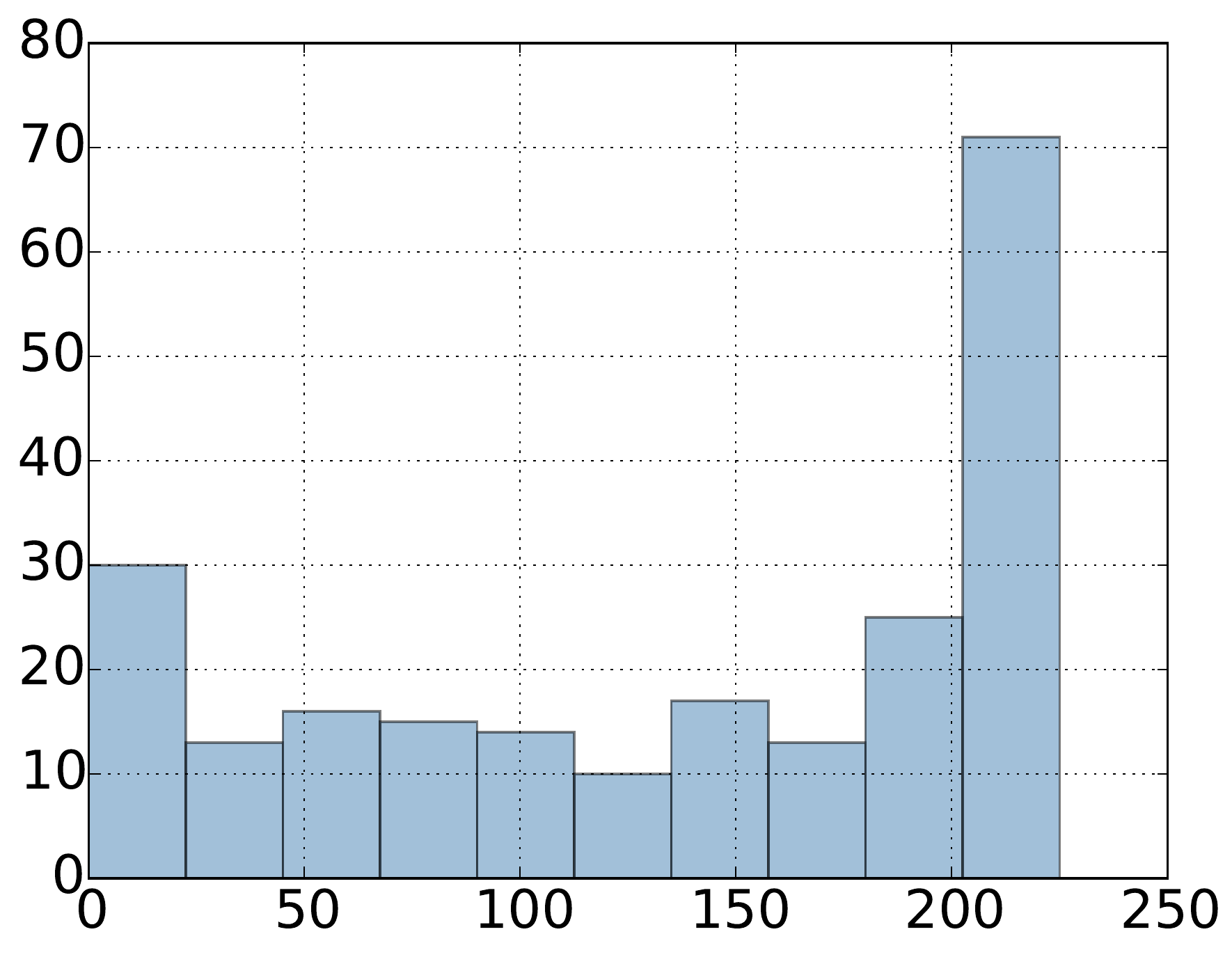} \\
  \raisebox{1.3em}{\rotatebox{90}{\scriptsize{Category}}}
	\includegraphics[width=0.19\linewidth]{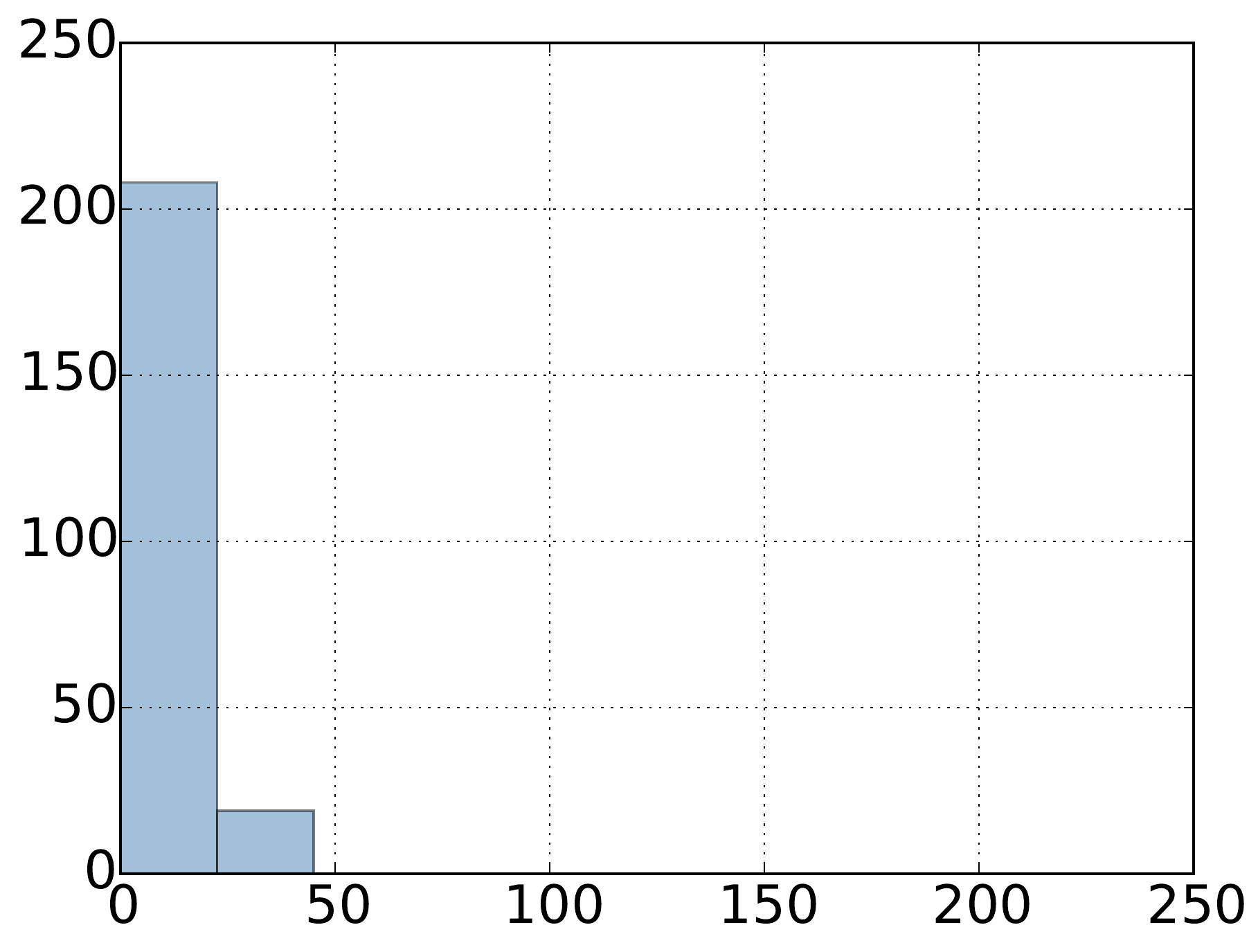} &
	\includegraphics[width=0.19\linewidth]{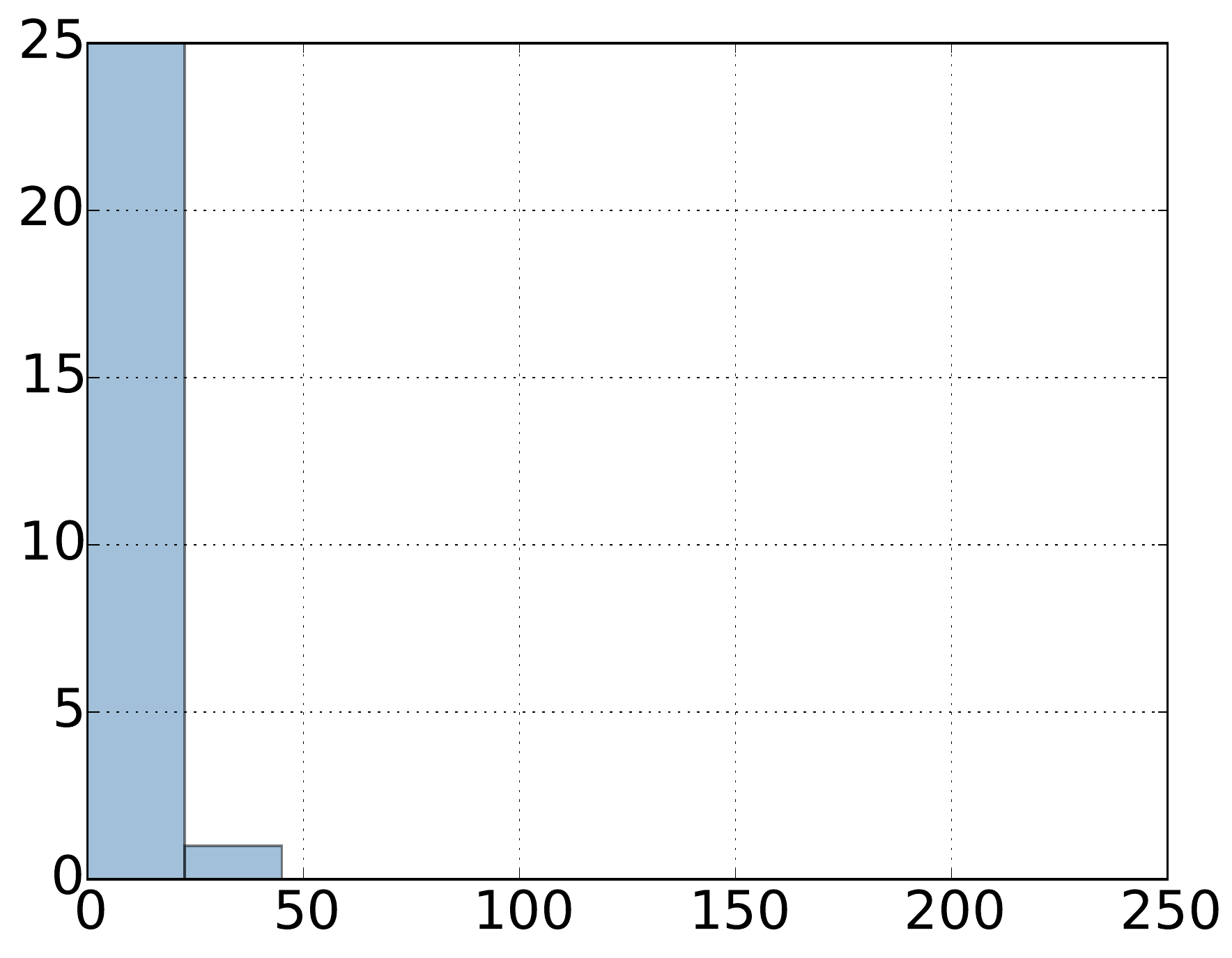} &
	\includegraphics[width=0.19\linewidth]{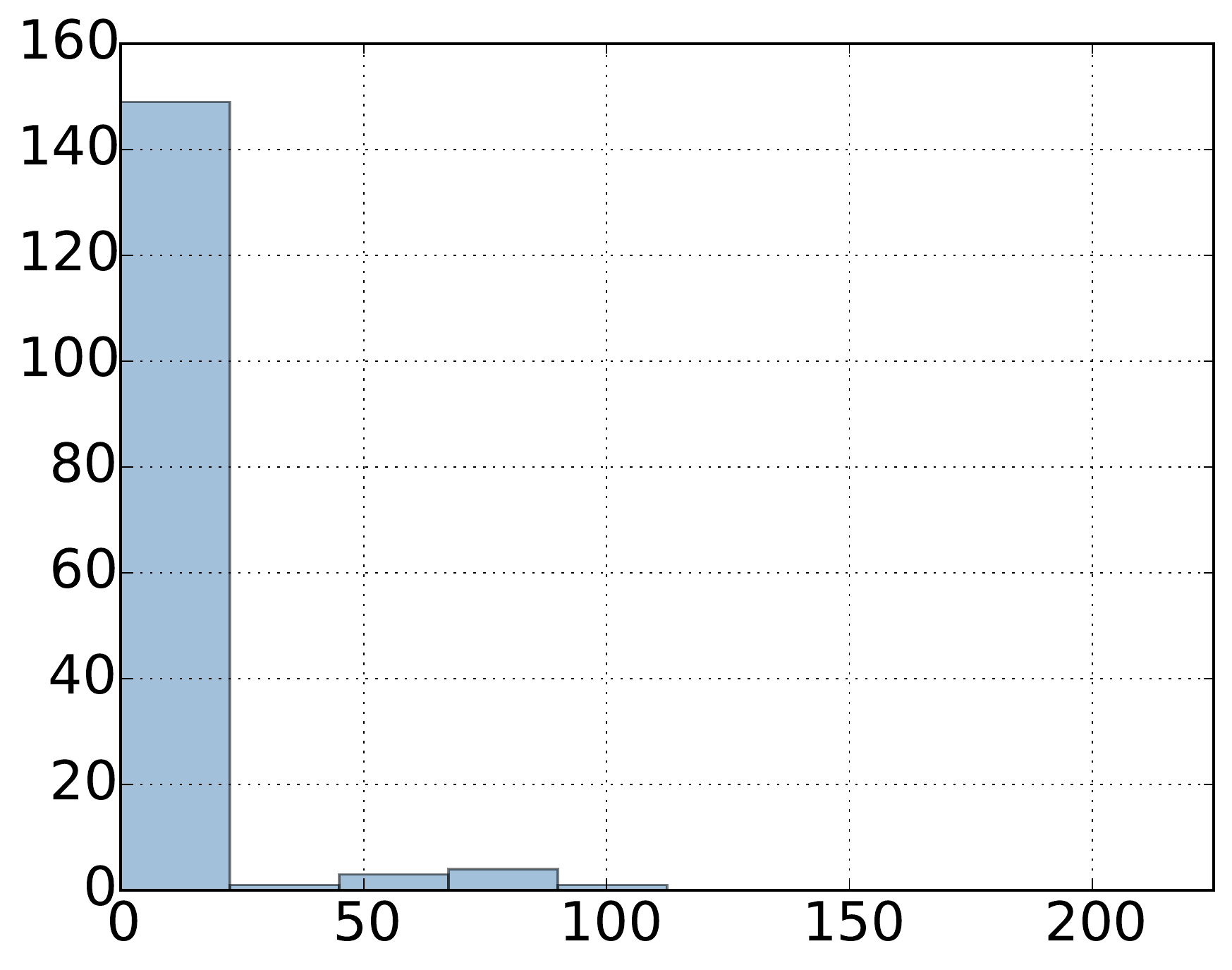} &
	\includegraphics[width=0.19\linewidth]{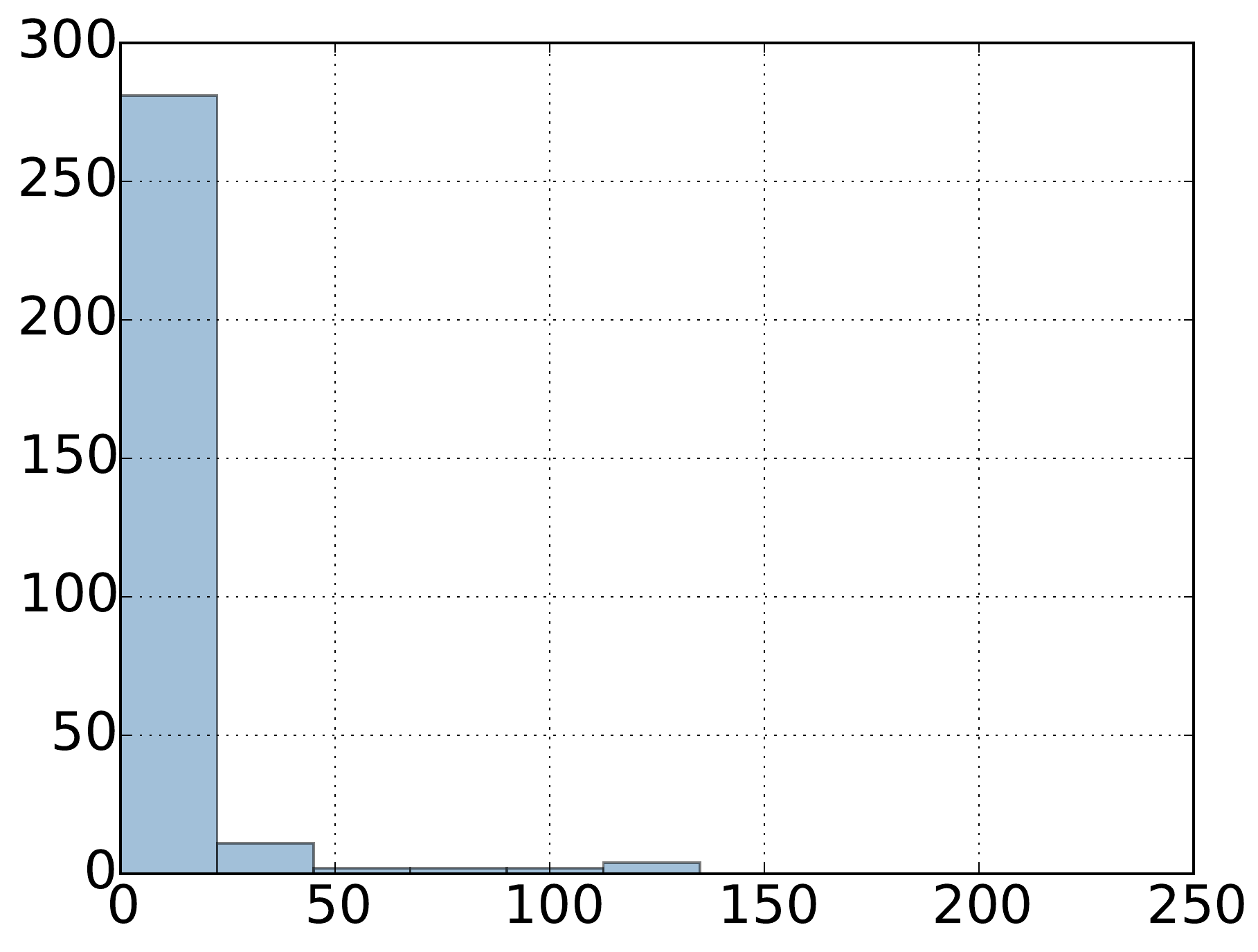} &
	\includegraphics[width=0.19\linewidth]{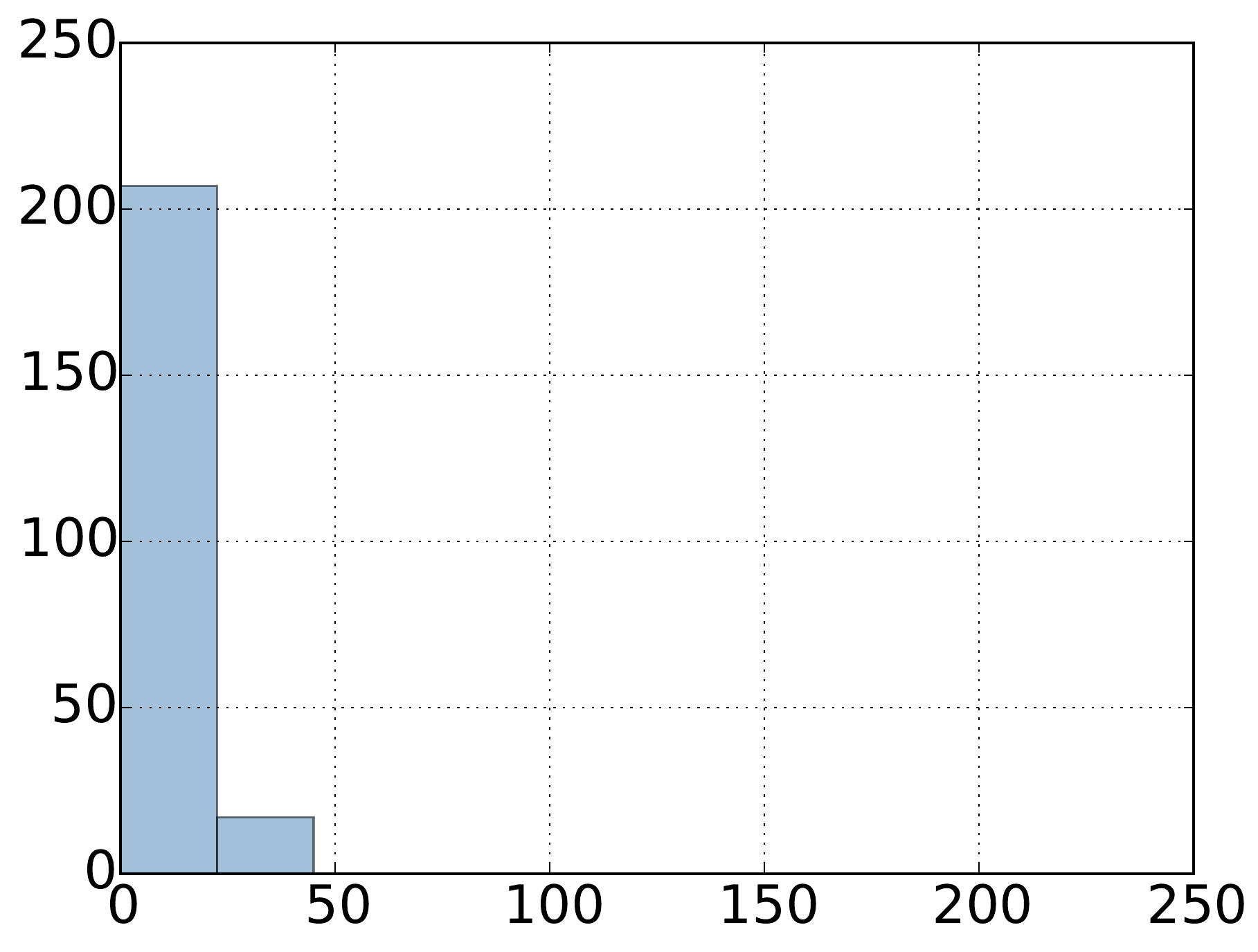} \\
\end{tabular}

\caption{Histograms over position in retrieval list obtained by our proposed approach 
(Y axis: \#images, X axis: position).
First row of histograms is over the position of instance match, and second is over position of category
match.}
\label{fig:class-hists}
\end{figure}

\begin{table}[t]
\centering
\caption{Mean recall @10 of ground truth model in retrievals for our method and baseline 
described in Sec.\ \ref{sec:cad_retrieval}}
\label{tab:IKEAImageNN-Quant}
\begin{tabular}{r@{~~~}ccccc@{~~~}c} \toprule
                  & Sofa    & Chair     & Bookcase  & Bed       & Table  & Overall\\ \midrule
Proposed          & \bf 32.3  & \bf 41.0    & \bf 26.8    & \bf 38.5    & 8.0     & \bf 29.3    \\
Fc7-NN            & 14.6      & 33.9        & 23.5        & 7.7        & \bf 17.4   & 19.4   \\ \bottomrule
\end{tabular} 
\end{table}

We now show results for retrieving CAD models from natural images. Our system
can naturally tackle this task: we map each model in the CAD corpus as
well as the
image to their latent representations, and perform a nearest neighbor
search in this embedding space.

We use cosine distance in the latent space for retrieval.
This approach is complementary to approaches like
\cite{Lim13,Lim14}: these approaches assume the existence of an exact-match 3D
model and fits the 3D model into the image. Our approach, on the other hand,
does not assume exact match and thus generalizes to retrieving the most similar object to
the depicted object (i.e., what is the next-most similar object in the
corpus). We show qualitative results in Fig.\ \ref{fig:cadRetrieval}.

We now quantitatively evaluate our approach. For each test window, we rank
all 225 CAD models in the corpus by cosine distance. We can
then determine 
two quantities: {\it (a) Instance match:} 
at what rank does the exact-match CAD model appear?
{\it (b) Category match:} at what rank does the first model of the same
category appear? 
As a baseline, we render all the 225 models at 30 deg.\ elevation and 8 uniformly sampled
azimuths from 0 to 360 deg.\ onto a white background, after scaling and translating
each model to a unit square at the origin. We then use ImageNet 
trained AlexNet's fc7 features over the query image and renderings to perform nearest neighbor
search (cosine distance). The first position at which a rendering of a model appears in the retrievals is taken
as the position for that model. Note that this is a strong baseline 
with access to lot more
information since it sees images, which are much higher resolution 
than our $20^3$ voxel grids. Moreover, it is significantly slower than our method, as it represents each 
3D model using 8 vectors of 4096D each, while our approach uses only a single 64D vector. 
As shown in Table\ \ref{tab:IKEAImageNN-Quant},
which reports the mean recall@10 of instance match,
we outperform
this baseline on all categories except tables/desks because 
most of the table models are very similar, and
fine differentiation between specific models
is very hard for a coarse $20^3$ voxel representation.
We report histograms of these ranks in Fig.\ 
\ref{fig:class-hists} per object category. 
For many categories, the 
top response is the correct category,
and the exact-match model is typically ranked highly.
Poor performance tends to result from images containing multiple objects
(e.g., a table picture with chairs in it), causing the network to predict
the representation for the ``wrong'' object out of the ambiguous input.
We also compare our model with \cite{Li15} in the supplement available on the project webpage.

\subsection{Shape Arithmetic}
\label{sec:exp_arithmetic}

\begin{figure}[t]
\includegraphics[width=\linewidth]{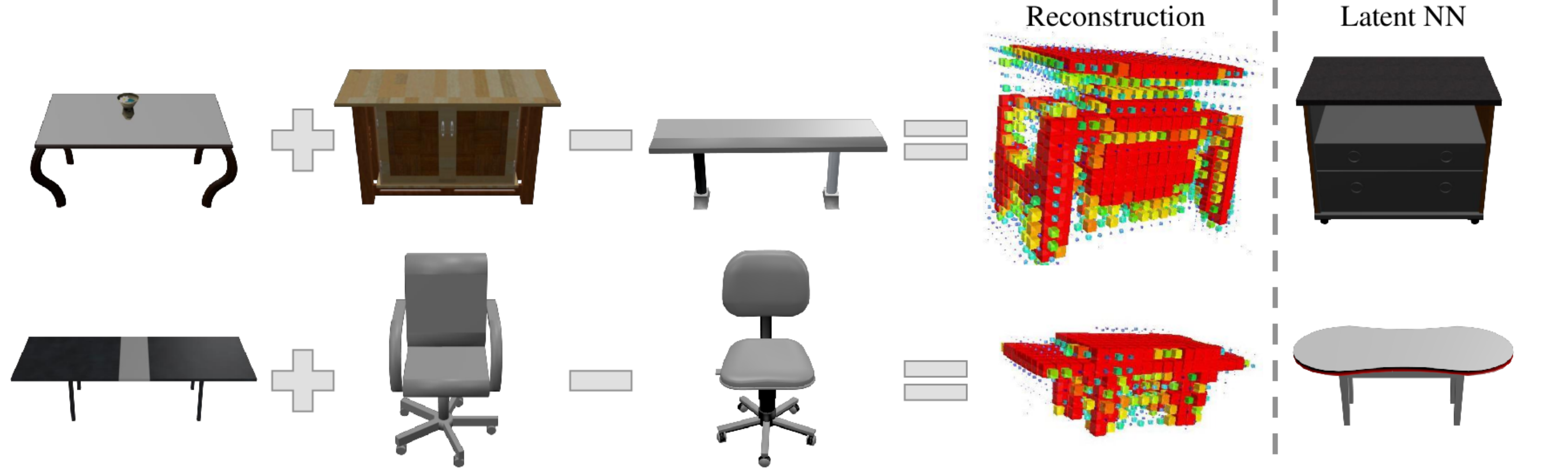} 
\caption{Results of shape arithmetic. In the first case, adding a cabinet-like-table to 
a table and removing small 2-leg table results in a table with built-in cabinet. In the second
case, adding and removing a similar looking chair with straight and curved edges respectively
leads to a table with curved edges.}
\label{fig:arithmetic}
\end{figure}

We have shown that the latent space is reconstructive and smooth, that it is
predictable, and that it carries class information. We now show some attempts
at probing the learned representation.
Previous work in vector embedding spaces~\cite{Radford_arxiv16,Mikolov_NIPS13} exhibit the phenomena of 
being able to perform arithmetic on these vector representations. For example, \cite{Mikolov_NIPS13} showed that
vector(”King”) - vector(”Man”) + vector(”Woman”) results in vector whose nearest neighbor was the 
vector for Queen.
We perform a similar experiment by randomly selecting 
triplets of 3D models and performing this $a+b-c$ operation on their latent representations. We then use the resulting 
feature to generate the voxel representation and also find the nearest neighbor in the dataset over 
cosine distance on this latent representation. We show some interesting
triplets in Fig.\ \ref{fig:arithmetic}.

{\small
\noindent \textbf{Acknowledgments:}
This work was partially supported by
Siebel Scholarship to RG, NDSEG Fellowship to DF and Bosch Young Faculty Fellowship to AG.
This material is based on research partially sponsored by ONR MURI N000141010934, ONR
MURI N000141612007, NSF1320083 and a gift from Google. The authors would like to thank Yahoo!
and Nvidia for the compute cluster and GPU donations respectively.
The authors would also like to thank Martial Hebert and Xiaolong Wang for many helpful discussions.
}

\clearpage

\bibliographystyle{splncs03}
\bibliography{egbib}



\section*{Supplementary material}

# Learning a Predictable and Generative Vector Representation for Objects
Supplementary Material

This is an abridged version with fewer and low-resolution results.
Full high-res version available on project webpage:
https://rohitgirdhar.github.io/GenerativePredictableVoxels/.

# Contents

# 1 Reconstruction Results on Synthetic Test Data

Some **randomly picked images** and corresponding reconstructions from the 23975 renderings of 4058 test models.

## 2 Reconstruction Results on Natural Images from IKEA Dataset

Select natural images and reconstructions on IKEA dataset.

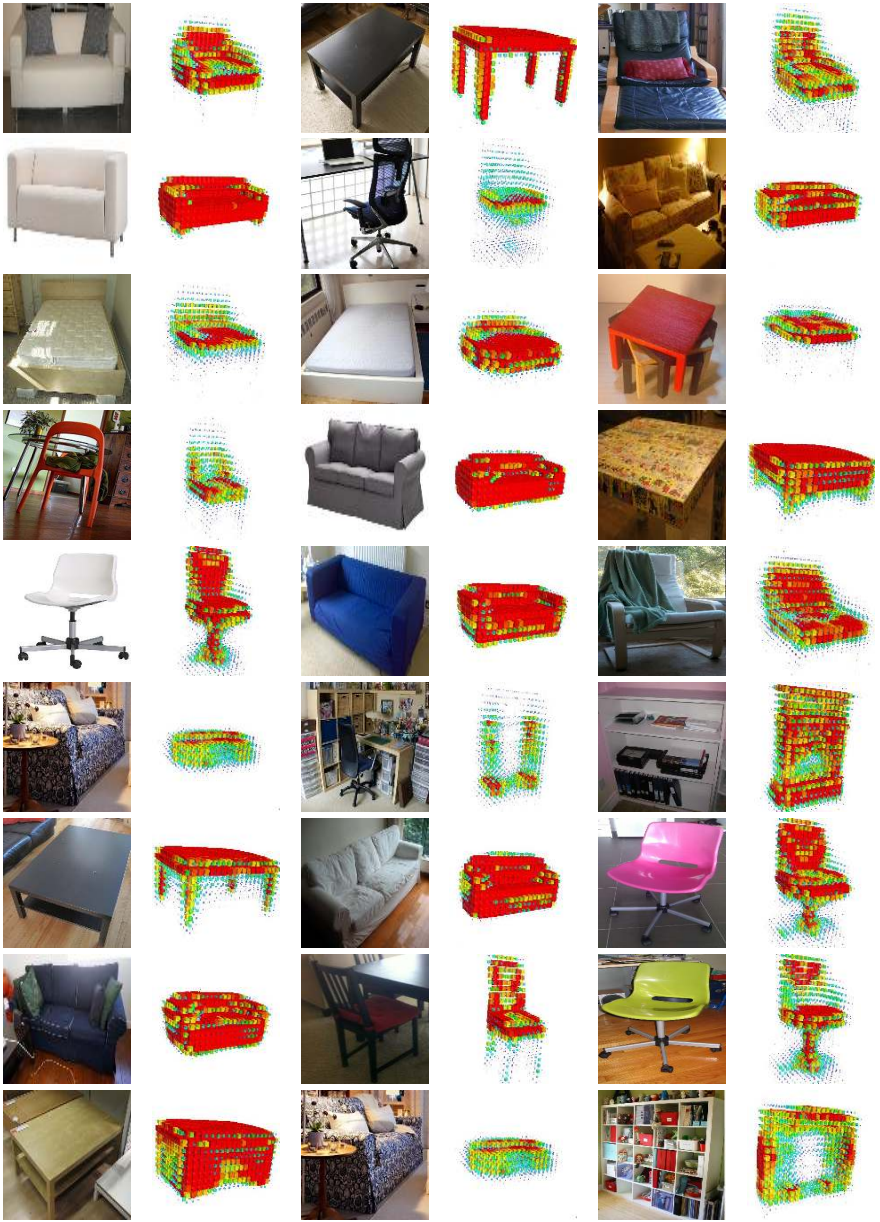

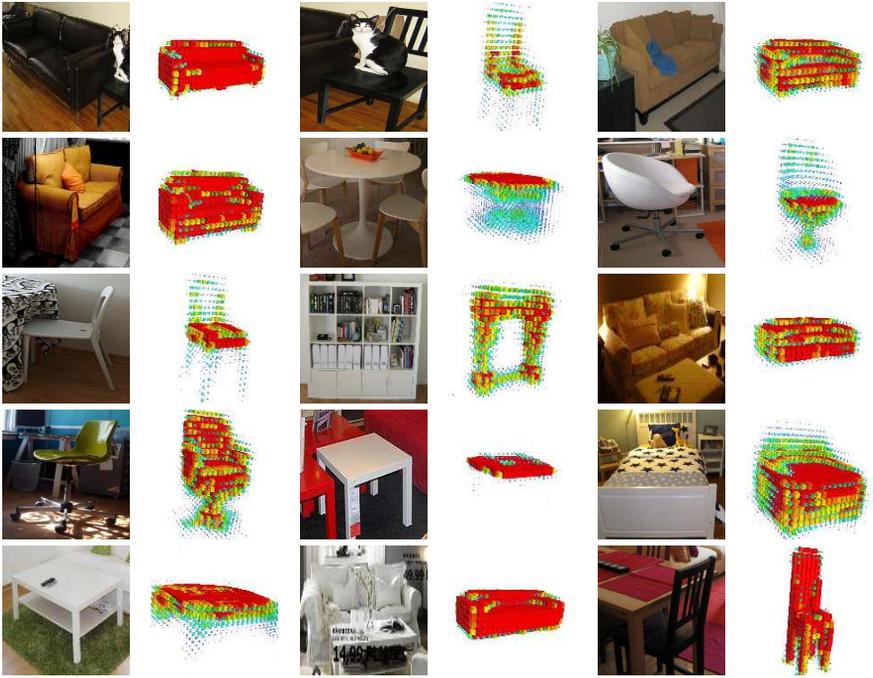

## 3 Nearest Neighbor on Natural Images from IKEA Dataset

Select natural images and 3D model nearest neighbors in IKEA dataset.

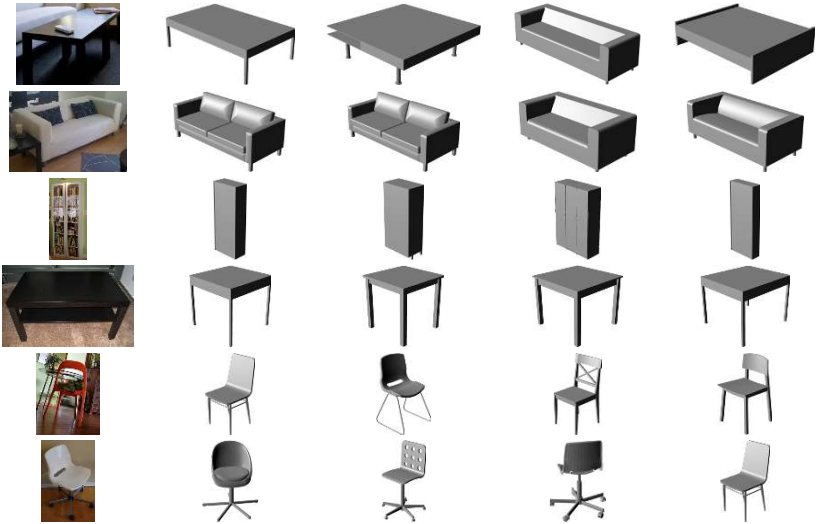

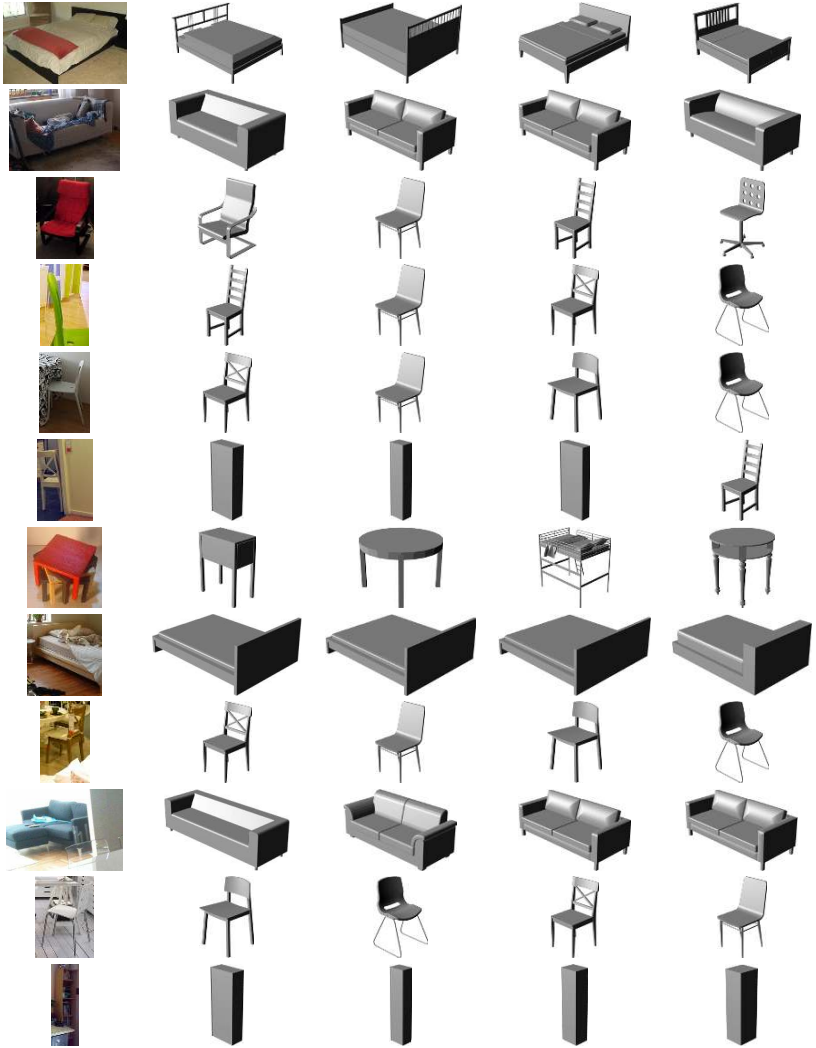

## 4 Comparison with Kar *et al.* [1] (3D Prediction)

### 4.1 Quantitative Evaluation

To compare, we first convert the prediction and ground truth obtained from the provided code by Kar *et al.* into an OBJ, by writing out the points and faces. Before computing the OBJ, we align the output points from their method with the ground truth as done in 'evaluation/evalMeshes.m' script provided by authors. We then voxelize the output to make it comparable with our approach.

## 4.2 Qualitative Results

Reconstructions on **randomly picked** Chair/Sofa images from PASCAL 3D+ using Kar *et al.* and our method. Complete qualitative results (on all 254 chair/sofa images) will be available on the project website.

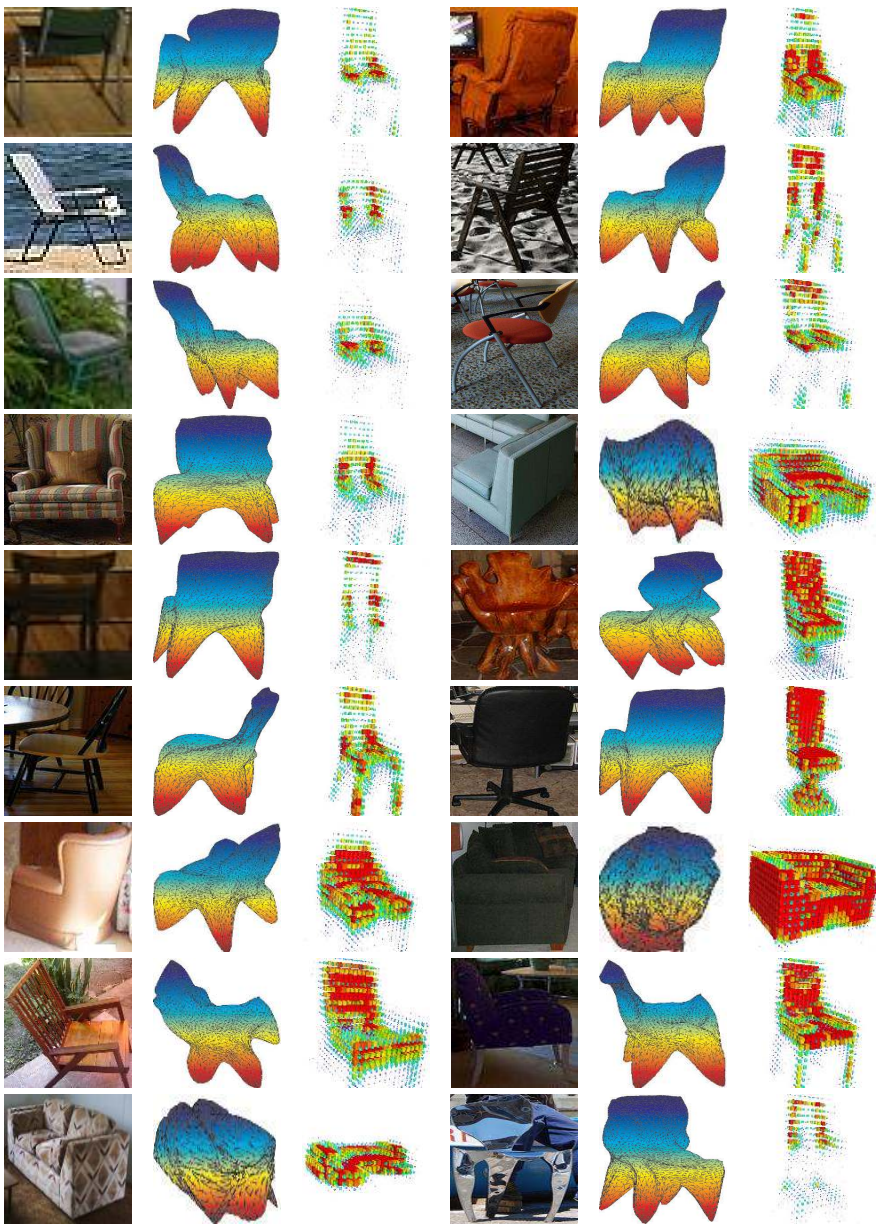

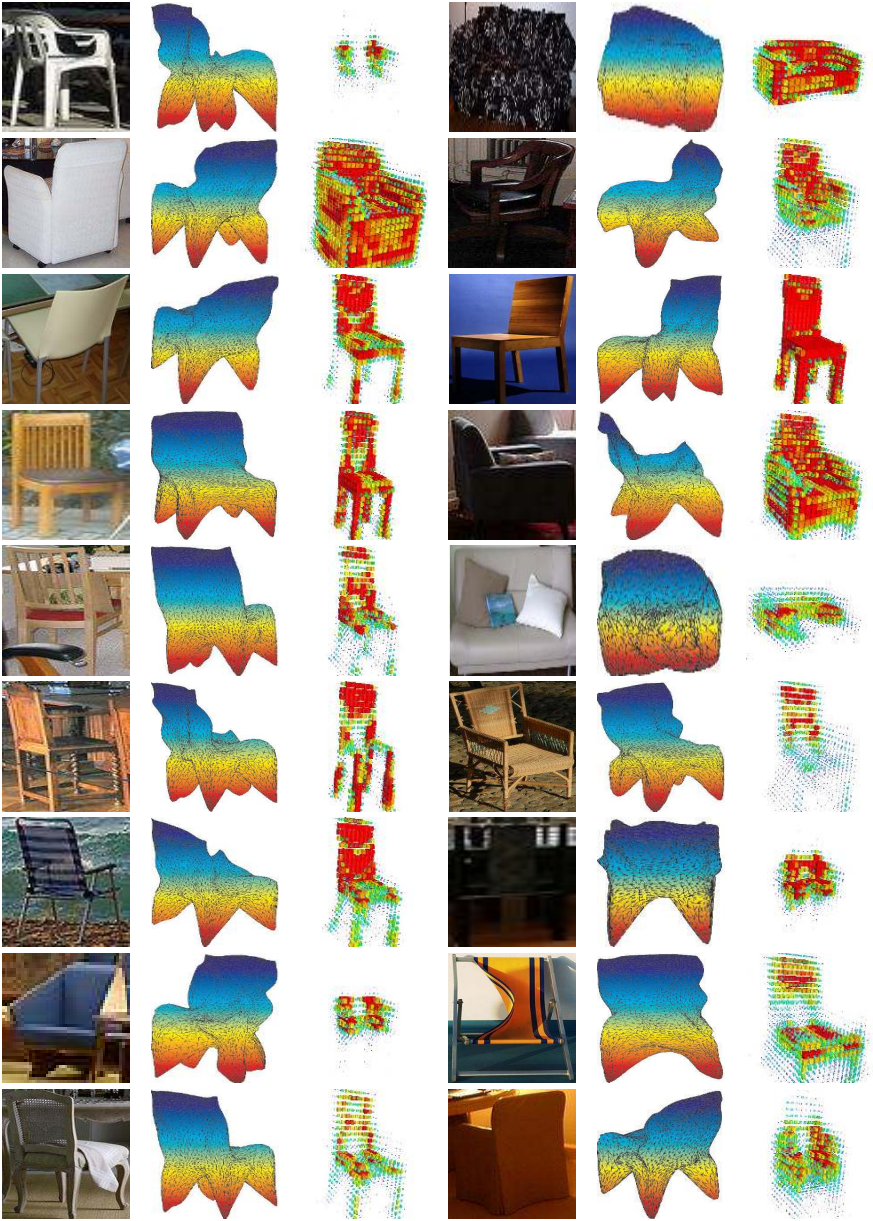

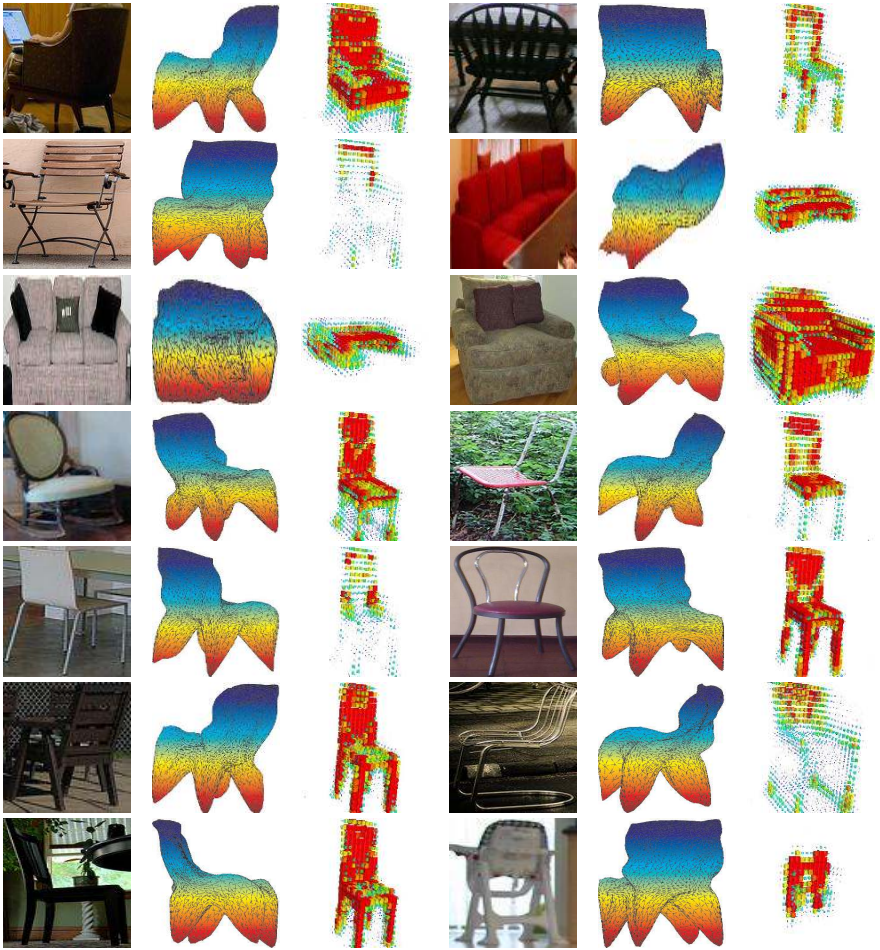

## 5 Comparison with Li *et al.* [2] (Image based 3D Model Nearest-neighbor search)

We now report a comparison with [2] on their 315 image, 105 model labeled evaluation set. [2]'s method is an approach that is specific to nearest-neighbor model retrieval and has a number of advantages over our approach. Their features are hand-crafted and extracted directly from the underlying 3D model, as opposed to learned automatically from a coarse $20^3$ voxel grid. Additionally, their method is designed to discriminate between similar models whereas our method's objective is trained purely for reconstruction. These two factors give Li et al.'s method substantial advantages in picking up on fine-grained details that would distinguish two chairs, for instance. As an added benefit, their approach is also class-specific. Despite these disadvantages, our method obtains strong performance. At recall@10, we get 82% compared to $\approx 95\%$ for them (obtained from Figure 8 in their paper).

8

# 6 More embedding space analysis

## 6.1 More interpolation Results

**Randomly picked results** for interpolation between two randomly picked models.

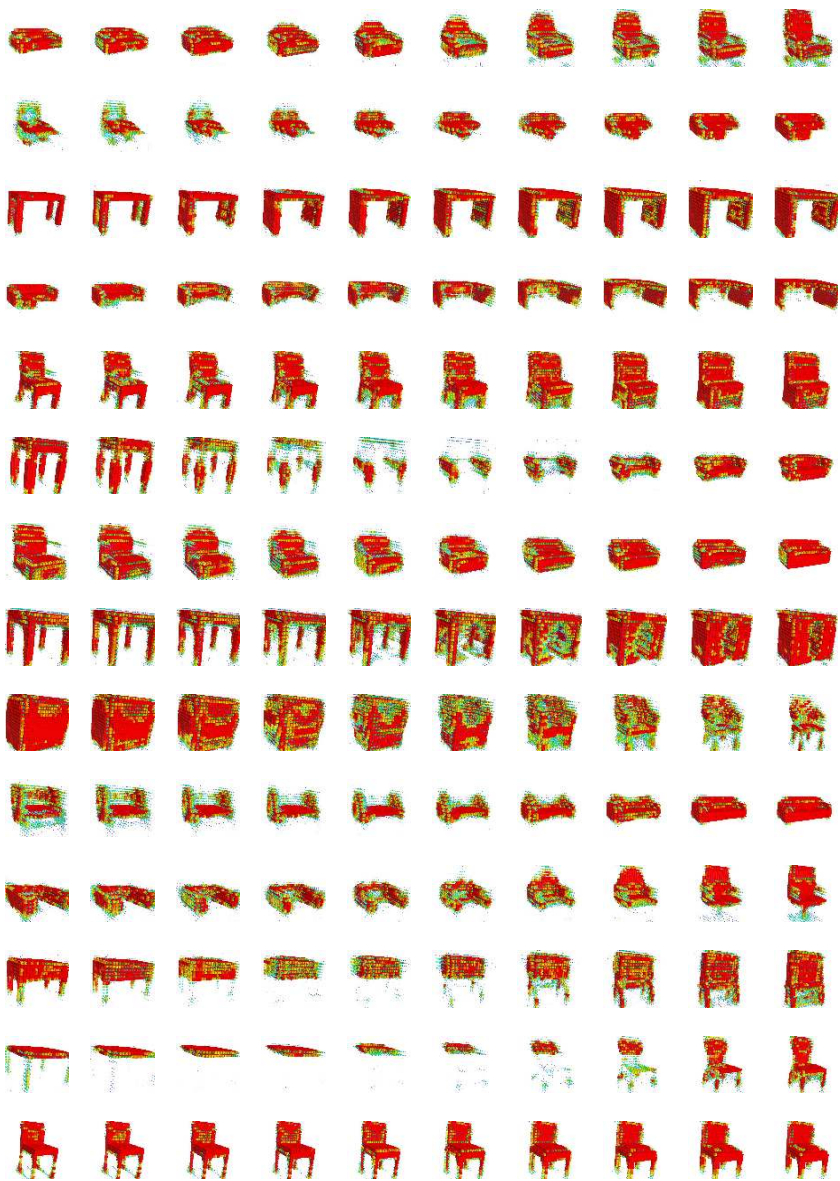

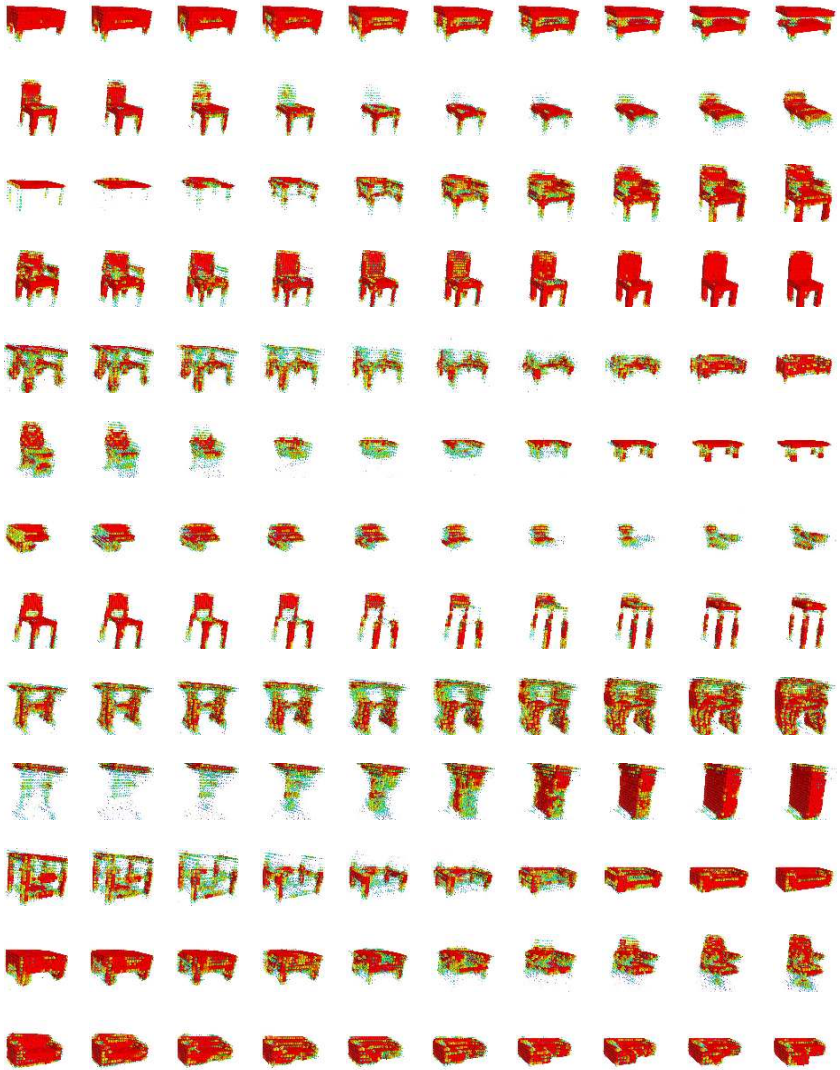



\end{document}